\documentclass{article}

   \usepackage[preprint, nonatbib]{neurips_2025}

\usepackage[utf8]{inputenc} 
\usepackage[T1]{fontenc}    
\usepackage{hyperref}       
\usepackage{url}            
\usepackage{booktabs}       
\usepackage{amsfonts}       
\usepackage{nicefrac}       
\usepackage{microtype}      
\usepackage{xcolor}         

\usepackage[utf8]{inputenc} 
\usepackage[T1]{fontenc}    
\usepackage{hyperref}       
\usepackage{url}            
\usepackage{booktabs}       
\usepackage{amsfonts}       
\usepackage{nicefrac}       
\usepackage{microtype}      
\usepackage{lipsum}
\usepackage{bm}
\usepackage{fancyhdr}       
\usepackage{graphicx}       
\graphicspath{{media/}}     
\usepackage{amsmath}
\usepackage{cleveref}
\usepackage{amssymb}
\usepackage{mathtools}
\usepackage{amsthm}
\theoremstyle{definition}

\usepackage{algorithm}
\usepackage{algorithmic}
\usepackage{booktabs}       
\usepackage{amsfonts}       
\usepackage{nicefrac}       
\usepackage{microtype}      
\usepackage{xcolor}  

\usepackage[utf8]{inputenc} 
\usepackage[T1]{fontenc}    
\usepackage{hyperref}       
\usepackage{url}            
\usepackage{booktabs}       
\usepackage{amsfonts}       
\usepackage{nicefrac}       
\usepackage{microtype}      
\usepackage{xcolor}         
\usepackage{pdfpages}
\usepackage{graphicx}

\usepackage{xspace}
\usepackage{array}
\usepackage{graphicx}
\usepackage{multirow,array}
\usepackage{multicol}
\usepackage[font=small]{caption}
\usepackage{subcaption}
\newtheorem{theorem}{Theorem}[section]

\newtheorem{lemma}[theorem]{Lemma}
\newtheorem{corollary}[theorem]{Corollary}

\usepackage{csquotes}
\usepackage{tcolorbox}
\usepackage{wrapfig}

\usepackage[T1]{fontenc}
\usepackage{enumitem}
\usepackage{listings}

\title{Exploration by Random Distribution Distillation}

\author{%
  Zhirui Fang\thanks{The first two authors contribute equally to this work.}\\
  Tsinghua University\\
  \texttt{fzr23@mails.tsinghua.edu.cn} \\
  \And
  Kai Yang$^*$ \\
  Tsinghua University \\
  \texttt{yk22@mails.tsinghua.edu.cn} \\
  \AND
  Jian Tao \\
  Tsinghua University \\
  \texttt{tj22@mails.tsinghua.edu.cn} \\
  \And
  Jiafei Lyu \\
  Tsinghua University \\
  \texttt{lvjf20@mails.tsinghua.edu.cn} \\
  \And
  Lusong Li \\
  JD Explore Academy \\
  \texttt{lilvsong@jd.com} \\
  \And
  Li Shen \\
  Sun Yat-sen University \\
  \texttt{shenli6@mail.sysu.edu.cn}
  \And
  Xiu Li \thanks{Corresponding author.}\\
  Tsinghua University \\
  \texttt{li.xiu@sz.tsinghua.edu.cn} \\
}

\PassOptionsToPackage{numbers, compress}{natbib}
\usepackage{amssymb}
\usepackage{multirow}

\begin{document}

\maketitle

\begin{abstract}
Exploration remains a critical challenge in online reinforcement learning, as an agent must effectively explore unknown environments to achieve high returns. Currently, the main exploration algorithms are primarily count-based methods and curiosity-based methods, with prediction-error methods being a prominent example. In this paper, we propose a novel method called \textbf{R}andom \textbf{D}istribution \textbf{D}istillation (RDD), which samples the output of a target network from a normal distribution. RDD facilitates a more extensive exploration by explicitly treating the difference between the prediction network and the target network as an intrinsic reward. Furthermore, by introducing randomness into the output of the target network for a given state and modeling it as a sample from a normal distribution, intrinsic rewards are bounded by two key components: a pseudo-count term ensuring proper exploration decay and a discrepancy term accounting for predictor convergence. We demonstrate that RDD effectively unifies both count-based and prediction-error approaches. It retains the advantages of prediction-error methods in high-dimensional spaces, while also implementing an intrinsic reward decay mode akin to the pseudo-count method. In the experimental section, RDD is compared with more advanced methods in a series of environments. Both theoretical analysis and experimental results confirm the effectiveness of our approach in improving online exploration for reinforcement learning tasks.
\end{abstract}
\section{Introduction}

Exploration remains a fundamental challenge in reinforcement learning, particularly in sparse-reward environments where extrinsic feedback is rare or delayed \cite{osband2016deep}. While count-based methods provide strong theoretical guarantees by incentivizing visitation to underexplored states \cite{strehl2008analysis,azar2017minimax}, their reliance on exact state counts limits their applicability to high-dimensional or continuous spaces \cite{sutton2022alberta}. Pseudo-count methods \cite{bellemare2016unifying,lobel2023flipping,ostrovski2017count,machado2020count} attempt to bridge this gap by approximating visitation statistics through density estimation, but they introduce additional complexity in training dynamics and hyperparameter sensitivity. Moreover, their theoretical connection to true counting relies on strong assumptions, such as stationary state distributions, which may not hold in practice \cite{ostrovski2017count}.  

Curiosity-driven exploration methods, such as those based on prediction error, offer greater scalability in complex environments \cite{pathak2017curiosity, burda2018exploration}. Approaches like Random Network Distillation (RND) \cite{burda2018exploration} leverage the mismatch between a fixed random target network and a learned predictor to generate intrinsic rewards. However, these methods lack a principled mechanism to ensure that the exploration bonus properly accounts for both state visitation frequency and the evolving relationship between the predictor and target networks \cite{strehl2008analysis,azar2017minimax,jin2018q}. This can lead to suboptimal exploration behavior in stochastic settings where the balance between novelty-seeking and reward exploitation is crucial.  

Recent advances, such as Distributional RND (DRND) \cite{yang2024exploration}, improve upon RND by aggregating bonuses from multiple random networks. While this provides better empirical performance, DRND still lacks a direct theoretical connection to state visitation counts and requires careful tuning of the number of target networks. The finite-sample approximation in DRND introduces practical limitations in how accurately it can model the underlying exploration dynamics. To address these limitations, we introduce Random Distribution Distillation (RDD), a novel exploration method that provides a unified framework connecting count-based principles with prediction-error approaches. RDD's intrinsic reward is naturally bounded by two key components: one that decreases with state visitation frequency similar to pseudo-count methods, and another that captures the discrepancy between the current predictor and its optimal version. This dual-term structure ensures that exploration bonuses properly decay with state familiarity while maintaining theoretical rigor. The RDD framework reveals important relationships with existing methods. When the target network's variance approaches zero, RDD reduces exactly to standard RND, showing that RND is a special case of our more general approach. Furthermore, as the number of target networks in DRND grows large, its behavior converges to RDD through the law of large numbers, establishing RDD as the asymptotic limit of DRND. These connections demonstrate how RDD unifies and extends previous exploration techniques while providing stronger theoretical guarantees.  

Our analysis shows that RDD's exploration bonus automatically adapts based on both state visitation counts and the learning progress of the predictor network. For frequently visited states, the bonus naturally decreases as both components decrease, while novel states receive proportionally larger bonuses. This adaptive behavior emerges directly from the RDD's theoretical foundations without requiring manual tuning, leading to more robust and efficient exploration across diverse environments. In the experiment, we compared existing online RL methods in diverse and rich environments to demonstrate the effectiveness of our approach. We also performed parameter ablation experiments and visualization experiments to further support our method.



\section{Related Work}
\label{gen_inst}
\textbf{Deep Reinforcement learning.}
Model-free reinforcement learning algorithms commonly employ variants of Q-learning \cite{watkins1992q} to learn the action-value function $Q_{\theta}(s,a)$, which is subsequently used to guide greedy action selection. This Q-function can be learned in high-dimensional state spaces through the use of non-linear function approximators parameterized by $\theta$. The learning process involves minimizing a loss function $L(\theta)=\mathbb{E}[(Q_{\theta}(s,a) - y_t) 
^2]$, where the target value $y_t$ for Q-learning is defined as \cite{mnih2015human}:
\begin{equation} \label{q}
y_{t} = 
{R}_{t}+\gamma \max_{a_{t+1}}Q_{\theta'}\left(s_{t+1},a_{t+1}\right) ,
\end{equation}
where $\theta'$ denotes the parameters of a slowly changing target network. This fundamental principle has been extended through various algorithmic advancements; for instance, Rainbow \cite{hessel2018rainbow} is tailored for discrete action spaces, while Soft Actor-Critic (SAC) \cite{haarnoja2018soft} is designed for continuous action spaces. Notably, most reinforcement learning systems rely on myopic exploration strategies—such as $\epsilon$-greedy or action noise—which are poorly suited for long-horizon tasks.

\subsection{Count-based exploration}
In tabular settings, count-based exploration rewards of the form $\frac{1}{\sqrt{N(s)}}$ (approximately) achieve a near-optimal balance between exploration and exploitation, even in highly stochastic Markov decision processes \cite{auer2002using,strehl2008analysis}. This approach has been extended to function approximation by employing density models to estimate pseudo-counts, initially using Context Tree Switching (CTS) models \cite{bellemare2016unifying} and subsequently with pixel convolutional neural networks \cite{ostrovski2017count}. However, learning density models in high-dimensional observation spaces remains challenging, particularly given the limitations discussed in the introduction. Subsequent Counting \cite{machado2020count} offers an alternative by relating pseudo-counts to the norm of successor representations \cite{dayan1993improving}, thus bypassing explicit density modeling. However, in tasks that require deep exploration \cite{taiga2021bonus}, these methods have been outperformed by pixel convolutional neural networks \cite{machado2020count}. Several methods circumvent the learning stage altogether, opting for heuristics that heavily incorporate domain knowledge, for example, $\#$-Counts \cite{tang2017exploration} and OPIQ \cite{initialisationoptimistic} utilize locality-sensitive hashing, Go-Explore \cite{ecoffet2021first} significantly reduces the sample size before binning input images, and MEGA posits prior knowledge about task-relevant state dimensions \cite{pitis2020maximum}. In contrast, CFN takes raw observations as input and flexibly learns optimized representations to predict exploration rewards.
\subsection{Model-prediction error exploration}
Several approaches involve learning a transition model and using the prediction error in the subsequent state as an intrinsic exploration reward \cite{stadie2015incentivizing,houthooft2016vime,pathak2017curiosity,ermolov2020latent,guo2022byol}. This encourages the agent to gather more data in areas where the transition model exhibits the highest inaccuracy \cite{kearns2002near,brafman2002r,kakade2003exploration}. However, most of these methods learn deterministic models \cite{pathak2017curiosity,guo2022byol}, which can limit their scalability to stochastic environments.
Due to its simplicity and empirical advantages, Random Network Distillation (RND) \cite{burda2018exploration} has emerged as a prominent exploration algorithm. RND assigns an intrinsic reward to a state based on a straightforward yet elegant heuristic: the novelty of a state is proportional to the discrepancy between the predictions of a trained network and a randomly initialized network for that state. In contrast to count-based methods, the exploration reward in RND lacks clear interpretability—it represents an unnormalized distance in a learned latent space rather than a visit-count statistic. Furthermore, the rate at which the RND reward diminishes with repeated visits is not well defined, as it depends on the learning dynamics of the neural network and stochastic gradient descent.

\section{Preliminaries}
\textbf{MDP.} We base our framework on the conventional Markov Decision Process (MDP) formulation as described in \cite{sutton1998introduction}. In this setting, an agent perceives the current state \( s \in \mathcal{S} \) and executes an action \( a \in \mathcal{A} \). The environment transitions to a new state \( s' \in \mathcal{S} \) according to the transition probability function \( P(s' \mid s, a) \), and the agent receives a reward \( r \) determined by the reward function \( r: \mathcal{S} \times \mathcal{A} \to \mathbb{R} \). The agent’s objective is to learn a policy \( \pi(a|s) \) that maximizes the expected cumulative discounted return:  $
\mathbb{E}_{\pi}\left[\sum_{t=0}^{\infty} \gamma^{t} r(s_t, a_t)\right],
$ where \( \gamma \in [0, 1) \) is the discount factor.

\textbf{Intrinsic reward.}
To enhance exploration, a common approach involves augmenting the agent's rewards in the environment with intrinsic rewards as a bonus. These intrinsic rewards, denoted as $b_t(s_t, a_t)$, incentives agents to explore unfamiliar states and take unfamiliar actions. 
Upon incorporating the intrinsic reward $b(s_t, a_t)$ into the original target of the Q value function, the adjusted target can be expressed as follows:
\begin{equation} \label{b}
r'_{t} = 
{r}_{t}+\lambda b\left(s_t, a_t\right) .
\end{equation}
Here $\lambda\in \mathbb{R}^+$ is the scale of the exploration bonus. 
Taking a greedy policy with respect to this Q-function can balance exploration and exploitation, forming the basis of numerous potentially efficient \cite{strehl2008analysis,jin2018q} and practically effective \cite{taiga2021bonus} algorithms. Consistent with common practice in the literature, we consider a state-dependent reward $b(s_t)$ \cite{burda2018large}.

\section{Methodology}
\label{headings}
\subsection{Random Distribution Distillation (RDD)}
Compared to RND, which distills a target network, we use the predictor network to distill the distribution of the target network. Given state $s$, suppose we have a random target distribution $\mathcal{N}(\mu_{\bar\theta}(s),\sigma_{\bar\phi}(s))$, where $\bar\theta$ and $\bar\phi$ are fixed and random parameters. Each time a target output $f_{\text{tar}}(s)$ is sampled from this distribution $f_{\text{tar}}(s) \sim \mathcal{N}(\mu_{\bar\theta}(s),\sigma^2_{\bar\phi}(s))$. We use the MSE loss to distill the target output by predictor network $f_{\theta}(s)$, i.e., minimize the loss below:
\begin{align}\label{loss}
    L(\theta) = \|f_{\theta}(s) - f_{\text{tar}}(s)\|^2, f_{\text{tar}}(s) \sim \mathcal{N}(\mu_{\bar\theta}(s),\sigma^2_{\bar\phi}(s)).
\end{align}
Assuming that the target network's output upon the $i$-th visit of the agent to state $s$ is $f_{\text{tar}}^i(s)$, we employ a prediction network to approximate this target network output, this optimal solution $f_{\theta}(s)$ of minimizing the $\mathrm{Eq.}$ \ref{loss} when visiting state $s$ for $n$ times and  is:
\begin{equation}
f^n_{\theta^*}(s) = \frac{1}{n} \sum_{i=1}^{n}f_{\text{tar}}^i(s), f_{\text{tar}}^i(s) \sim \mathcal{N}(\mu_{\bar\theta}(s),\sigma^2_{\bar\phi}(s)).
\end{equation}

In the context of RND, as the prediction network is trained more frequently on the same state, its output converges towards the target network's output. Consequently, directly using the prediction loss as a reward incentivizes the agent to explore novel states. However, our method does not equate the reward with the loss, as the loss associated with fitting a random variable can be unstable. The expected value of the prediction network is given by:
\begin{equation} \label{mean}
\mathbb{E}\left[f^n_{\theta^*}(s)\right] = \mathbb{E}\left[\frac{1}{n} \sum_{i=1}^{n}f_{\text{tar}}^i(s)\right] = \mu_{\bar\theta}(s).
\end{equation}

Tracking the frequency of data occurrences is crucial for addressing inconsistencies in terminal rewards. Traditional count-based methods rely on large tables to record state visit counts, whereas pseudo-count strategies leverage neural networks for estimation, offering a scalable perspective on state visitation. Nevertheless, these approaches can introduce computational and storage complexities, particularly when dealing with high-dimensional inputs. We construct a statistic 
$z_n(s)$ capable of indirectly estimating state occurrences without the need for auxiliary functions:
\begin{equation}
z_n(s) = \frac{\|f^n_{\theta^*}(s)- \mu_{\bar\theta}(s)\|^2}{\sigma^2_{\bar\phi}(s)}.
\end{equation}
The statistic $z_n(s)$ converges unbiasedly to the reciprocal of state visitation counts $1/n$, while maintaining exceptionally low estimation variance.

\subsection{Theoretical Analysis}
\label{Theoretical}
In this section, we will compare the statistics $z_n(s)$ with $1/n$ provided by the count-based method, as well as the statistics $y_n(s)$ given by DRND, analyze their unbiasedness, effectiveness and consistency.

\begin{lemma}\label{Lemma1}
$f^n_{\theta^*}(s)$ is the optimal function of $\mathrm{Eq.}$ \ref{loss} when RL agent visits the state $s$ for $n$ times, the statistic
\begin{equation} \label{statistic}
z_n(s) = \frac{\|f^n_{\theta^*}(s)- \mu_{\bar\theta}(s)\|^2}{\sigma^2_{\bar\phi}(s)}.
\end{equation}
is an \textbf{consistent and unbiased estimator} of $1/n$.
\end{lemma}

The complete proof is provided in Appendix \ref{proof_A} and Theorem \ref{consistency}. Lemma \ref{Lemma1} establishes that the statistic $z_n(s)$ provides an unbiased estimate of the reciprocal state visitation count $1/n$. The following corollary further demonstrates that $z_n(s)$ exhibits superior stability compared to DRND's $y_n(s)$.

\begin{corollary}\label{coro}
  When the target network $f_{\text{tar}}^i(s) \sim \mathcal{N}(\mu_{\bar\theta}(s),\sigma^2_{\bar\phi}(s)$), the estimator \( z_n(s) \) maintains the same expectation as DRND's $y_n(s)=\sqrt{\frac{[f_{\theta}(x)]^2-[\mu(x)]^2}{\frac{1}{N}\Sigma_{i=1}^{N}\left(\bar{f}_i(x)\right)^2-[\mu(x)]^2}}$ ($\bar{f}_i(x)$ represents the $i$-th target network), i.e.,  
\[ \mathbb{E}[z_n(s)] = \mathbb{E}[y_n(s)] = 1/n, \]  
while achieving strict variance reduction:  
\[\mathrm{Var}(z_n(s)) \leq \mathrm{Var}(y_n(s)). \]  
\end{corollary}

The complete proof is provided in Appendix \ref{proof_B}. This implies that both $y_n(s)$ and $z_n(s)$ exhibit \textbf{unbiased} statistical properties. From the perspective of variance, the statistic $z_n(s)$ proposed in this paper has a smaller variance compared to the $y_n(s)$ proposed by the DRND method, providing a better and robust estimate for $\frac{1}{n}$. This indicates that our statistical method is more \textbf{effective} than $y_n(s)$ in DRND.

The preceding Lemma \ref{Lemma1} establishes the convergence result but does not characterize its rate. The following theorem quantifies the convergence rate at which the statistic $z_n(s)$.

\begin{theorem}\label{consistency}
    The expression 
    \[
    z_n = \frac{\|f^n_{\theta^*}(s) - \mu_{\bar \theta}(s)\|}{\sigma^2_{\bar \phi}(s)}
    \]
    satisfies the following inequality:
    \[
    \mathbb{P}\left( \left| z_n - \frac{1}{n} \right| < \epsilon \right) > 1 - 2\delta,
    \]
 where $\epsilon=\sqrt{\frac{2}{\delta n^3}}+\frac{ \sqrt{ \ln\left( \frac{2}{\delta} \right) \left[ \frac{n(n-1)}{2} + \ln\left( \frac{2}{\delta} \right) \right] } }{n^2 }$.
 \end{theorem}
 The complete proof is provided in Appendix \ref{proof_C}. $z_n(s)$ exhibits a convergence rate ranging from $O(n^{-1})$ to $O(n^{-1.5}$), with guaranteed convergence to the expected value in the limit. For $\forall\epsilon>0,\ \lim_{n\rightarrow+\infty}\mathbb{P}(|z_n-\frac{1}{n}|<\epsilon)=0$. It has been demonstrated that $z_n(s)$ is equipped with \textbf{consistency}.

\subsection{Final bonus of RDD agent}

In the actual implementation of the algorithm, we did not assign a unique variance to each distribution; instead, we set them all to a fixed value of $\sigma^2$. This approach helps reduce the number of parameters and prevents overly dispersed sampling caused by excessively large variances, which can lead to convergence issues in the predictor network. Moreover, in the actual implementation, we referenced the CFN \cite{lobel2023flipping} method. We will set the predictor network to be $d$-dimensional, aligning with the dimensions of the target network, and then update using MSE loss. Setting the dimensions to $d$-dimensional will not affect the expectation but can reduce the variance of the statistical quantity $z_n$. The specific analysis is as follows:
\begin{align}
\bm{f}^n_{\theta^*}(s) &= \frac{1}{n} \sum_{i=1}^{n}\bm{f}_{\text{tar}}^i(s), \quad \bm{f}_{\text{tar}}^i(s) \sim \mathcal{N}(\bm{\mu}_{\bar\theta}(s) = \mu_{\bar\theta}(s) \times \mathbf{1}_d,\mathbf{\sigma^2} \times \mathbf{1}_d).
\end{align}
\begin{align}\label{E}
\mathbb{E}\left[\frac{\|\bm{f}^n_{\theta^*}(s)- \bm{\mu}_{\bar\theta}(s)\|^2}{d{\sigma^2}} \right] &= \mathbb{E}\left[\frac{1}{d{\sigma^2}}\sum_{j=1}^d\frac{1}{n}\sum_{i=1}^{n}\|f_{\text{tar}}^{i,j}(s)-\mu_{\bar{\theta}}(s)\|^2\right], f_{\text{tar}}^{i,j}(s) \sim \mathcal{N}(\mu_{\bar\theta}(s),{\sigma^2}) \nonumber\\
&= \frac{1}{d}\mathbb{E}[\sum_{j=1}^d z^j_n] = \frac{1}{d}\mathbb{E}[\sum_{j=1}^d z_n] = \frac{1}{n}.\\
\text{Var}\left[\frac{\|\bm{f}^n_{\theta^*}(s)- \bm{\mu}_{\bar\theta}(s)\|^2}{d{\sigma^2}} \right] &= \text{Var}\left[\frac{1}{d{\sigma^4}}\sum_{j=1}^d\left[\frac{1}{n}\sum_{i=1}^{n}\|f_{\text{tar}}^{i,j}(s)-\mu_{\bar{\theta}}(s)\|^2\right]  \right] \nonumber\\&= \frac{1}{d^2}\text{Var}[\sum_{j=1}^d z_n] = \frac{1}{d}\text{Var}[z_n].
\end{align}
Using this method, it is possible to reduce the variance of $z_n$, resulting in estimates that are closer to the true values, thus improving the accuracy of the prediction. Furthermore, this approach only requires modifying the dimensionality of network outputs, without introducing significant computational overhead. Based on the analysis in the previous section, we can modify the RDD bonus to the following final form, which is the encouragement bonus used in our algorithm to incentivize agent exploration:
\begin{align}\label{bonus}
b(s) = \frac{1}{d}\|\bm{f}_{\theta}(s)- \bm{\mu}_{\bar\theta}(s)\|^2.
\end{align}
Through the final reward, we can obtain algorithm process and pseudo-code in Appendix \ref{Diagram} and \ref{Pseudo-code}.
\subsection{Relationship between Pseudo-count, RND, DRND and RDD}
\label{Relationship}
\textbf{Relationship of pseudo-count methods.}
Let's rewrite $\mathrm{Eq.}$\eqref{E} as follows:
\begin{align}
\mathbb{E}\left[\|f^n_{\theta^*}(s)- \mu_{\bar\theta}(s)\|^2\right] = \frac{\sigma^2_{\bar\phi}(s)}{n}.
\end{align}
Please note that the expectation is taken over samples from the target network $f_{\text{tar}}$, therefore, $f_{\theta}(s)$ and $\mu_{\bar\theta}(s)$ can be treated as constants when taking the expectation. By using triangle inequality, we can obtain the following expression:
\begin{align}
\|f_{\theta}(s)- \mu_{\bar\theta}(s)\|^2 = \mathbb{E}[\|f_{\theta}(s)- \mu_{\bar\theta}(s)\|^2] &\le \mathbb{E}[\|f^n_{\theta^*}(s)- \mu_{\bar\theta}(s)\|^2] + \mathbb{E}[\|f^n_{\theta^*}(s)- f_{\theta}(s)\|^2] \nonumber\\
& =  \underbrace{\frac{\sigma^2_{\bar\phi}(s)}{n}}_\text{pseudo-count term}+ \underbrace{\mathbb{E}[\|f^n_{\theta^*}(s)- f_{\theta}(s)\|^2]}_\text{discrepancy term}.
\end{align}

Here $f_{\theta}(s)$ represents the predictor network with current parameters, $f^n_{\theta^*}(s)$ denotes the optimal parameter network after $n$ visits to state $s$, and $\mu_{\bar\theta}(s)$ and $\sigma^2_{\bar\phi}(s)$ are known fixed parameters of the target network distribution. We can use $\|f_{\theta}(s)- \mu_{\bar\theta}(s)\|^2$ as the RDD agent bonus, which is similar but a little different from the RND agent, since the RDD agent uses the mean value of all output from the target network. This bonus is the lower bound of the formula on the right. The expression on the right consists of two terms: Since $\sigma^2_{\bar\phi}(s)$ is a constant, the first pseudo-count term $\sigma^2_{\bar\phi}(s)/n$ is $1/n$ times a constant; the discrepancy term represents the discrepancy between the optimal predictor and the current predictor. The current bonus is constrained by the number of visits and the difference between the current predictor and the optimal predictor. When the state $s$ is visited frequently, both factors decrease, constraining the lower bound of the bonus, resulting in a smaller bonus value. Conversely, for states with fewer visits, the lower bound is wider, allowing for a larger bonus value.

\textbf{Relationship of RND methods.} It is worth noting that since the variance of the target network is given by predefined random parameters, if we set it to a small value approaching zero, the distribution of the target network becomes an approximate delta distribution, where all target network samples $f_\text{tar}(s) \sim \mathcal{N}(\mu_{\bar\theta}(s),\sigma^2_{\bar\phi}(s)\approx 0)$ are almost equal to the mean. At this point, both the bonus and loss expressions have degenerated to be the same as RND, and the method has also degenerated to RND. In other words, RND is a special case of RDD, while RDD is an extension of the RND method.

\textbf{Relationship of DRND methods.} In DRND, the estimation of state visitation counts is derived under the premise of a finite number of target networks, with known mean and variance outputs from those networks. In contrast, RDD approximates state visitation counts assuming a normal distribution with known parameters. As the number of target networks in DRND increases, according to the law of large numbers, the resulting distribution will converge to a normal distribution, effectively transforming into the RDD method. Therefore, it can be considered that the target networks in DRND are sampled from RDD, and RDD can be viewed as the DRND method with an infinite number of target networks.

\section{Experiments}
\label{experi}
In this section, we first present a numerical example to support our viewpoints in Section \ref{Theoretical} and Section \ref{Relationship}. Subsequently, we test the superiority of our method over other baseline methods in well-known environments, and in Table \ref{tabel}, we have listed the episode returns of these algorithms. Then, in the MountainCar simulation environment, we investigate the exploration process by visualizing the probability density of the states. Finally, we investigate the hyperparameter settings of the method.
\subsection{Numerical Example}
In this section, we present our numerical experiments to support some of the statistical properties of our method. We sampled 100 points from the replay buffer of the online reinforcement learning experiment and used t-SNE to reduce the dimension into a two-dimensional space, thereby forming a mini-dataset. We then trained the RND, DRND and RDD models on this mini-dataset. We represent the statistics $y_n(s)$ and $z_n(s)$ in DRND and RDD based on the exact values of the samples accessed currently as $y$ and $z$. The predicted values obtained by the predictor networks are called $y_{approx}$ and $z_{approx}$. The output bonus of the RND network for the current sample is represented as $rnd$, the rewards obtained through the counting method are represented as $1/n$. 

\begin{figure}[htp]
  \centering
  \begin{subfigure}{0.32\textwidth}
    \centering
    \includegraphics[width=1\linewidth]{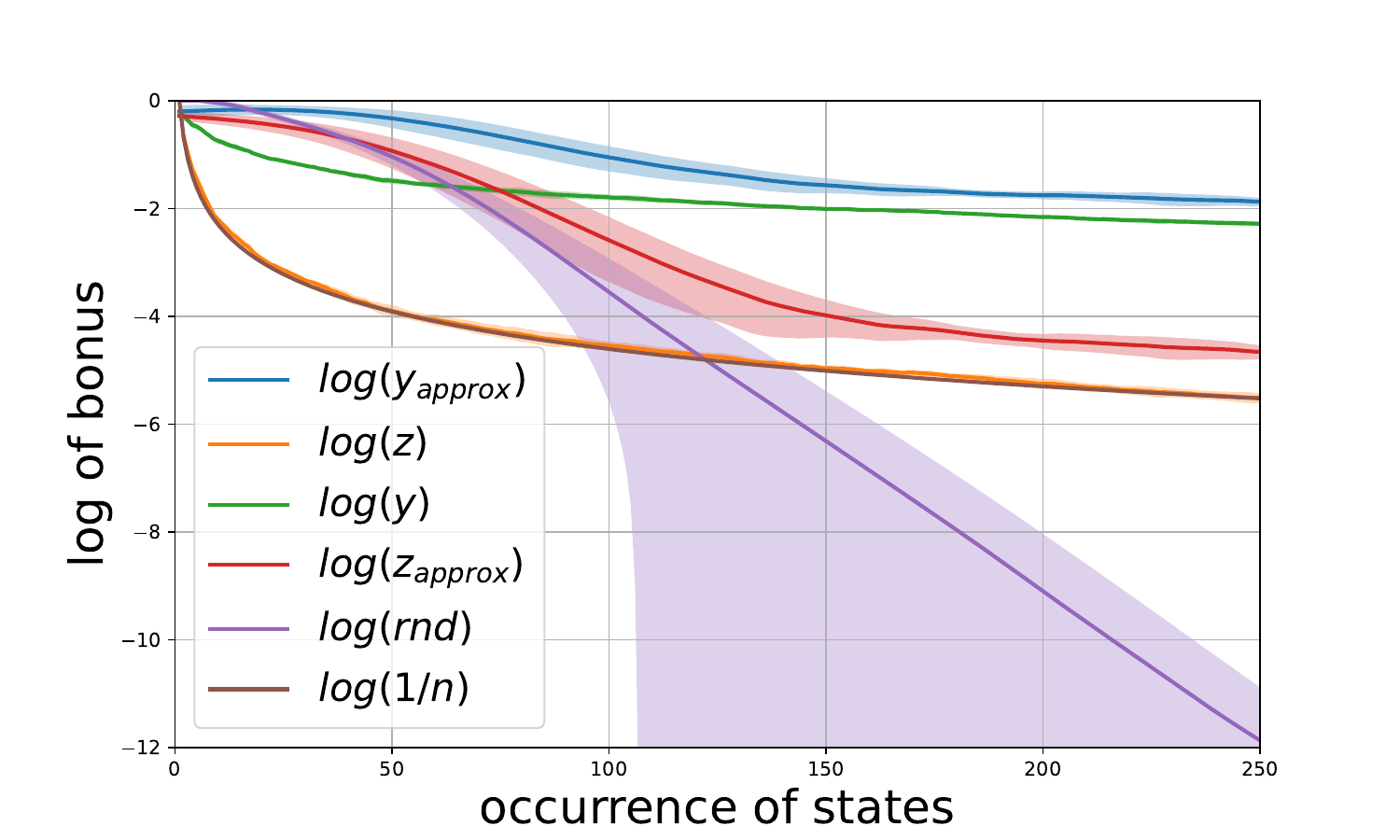}
    \captionsetup{font=scriptsize}
    \caption{} 
    \label{toy1}
  \end{subfigure}
  \begin{subfigure}{0.33\textwidth}
    \centering
    \includegraphics[width=1\linewidth]{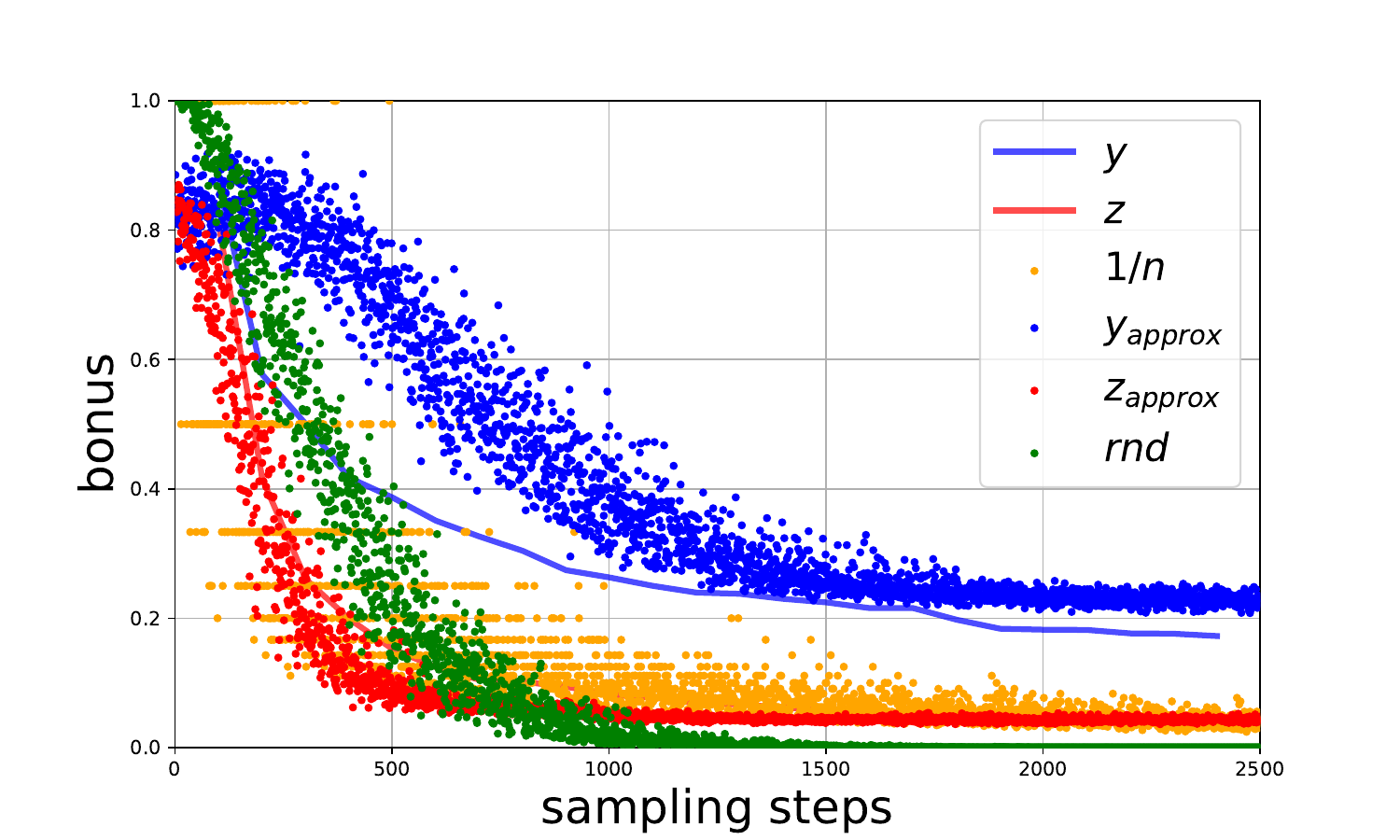}
    \captionsetup{font=scriptsize}
    \caption{} 
    \label{toy2}
  \end{subfigure}
  \begin{subfigure}{0.33\textwidth}
    \centering
    \includegraphics[width=1\linewidth]{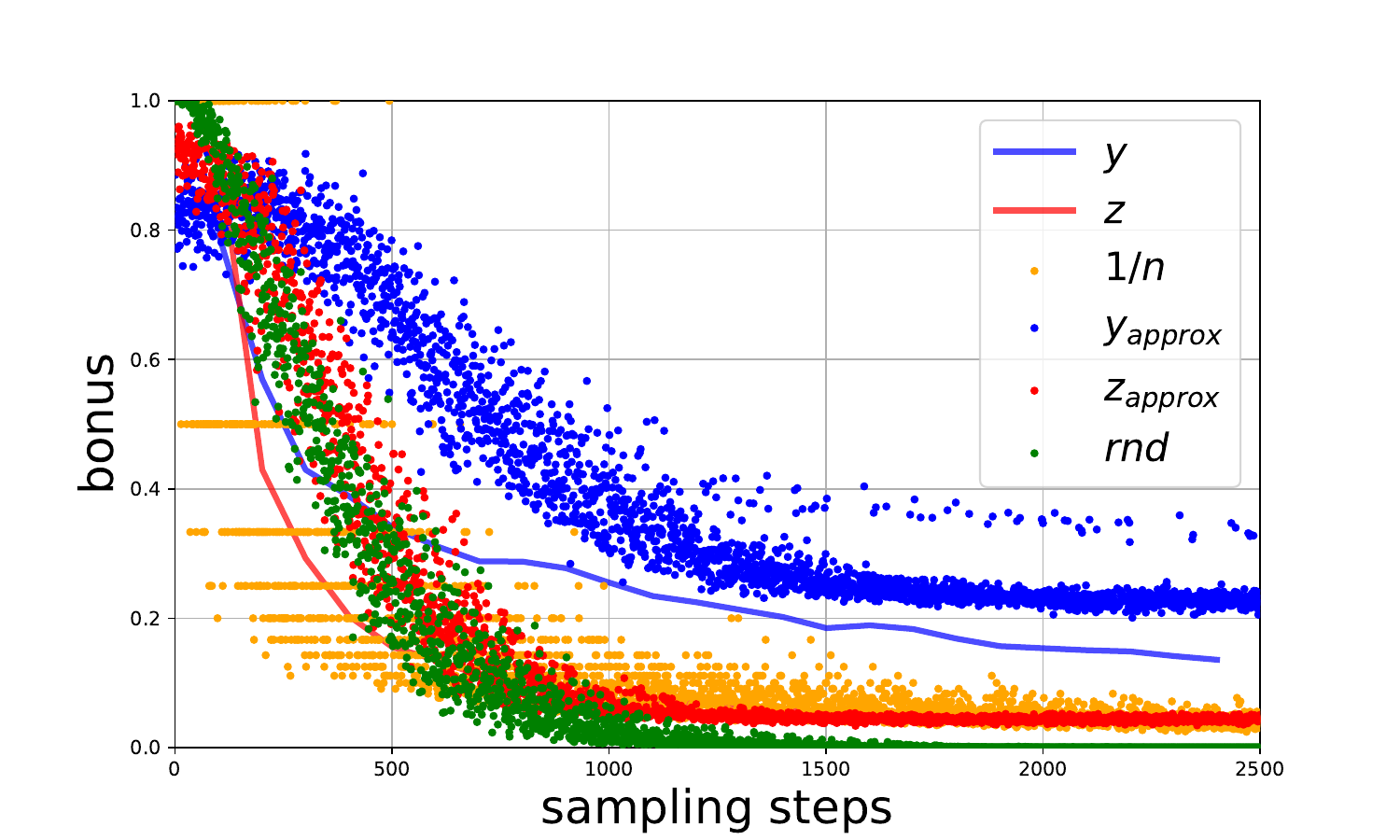} 
    \captionsetup{font=scriptsize}
    \caption{} 
    \label{toy3}
  \end{subfigure}
\caption{Numerical Example. We investigated the decay of bonus with each access to the same state when the bonus remains constant, as well as the bonus obtained by the agent through random wandering (simulating a random strategy) on the mini-dataset.}
\end{figure}

In Figure \ref{toy1}, we sampled 5 points from the mini-dataset to study the decay of bonus as the number of visits to the same state (those 5 points) increases. We initialized target networks for RND, DRND, RDD to $f_\text{tar}(s) \sim \mathcal{N}(\mu\times\bm{1_d},\sigma^2\times\bm{1_d}), \mu=1, \sigma=1, \bm{d}=256.$ Based on the results of the reward decay, rnd shows a significant difference in the rewards for different states with the same number of visits. The statistics $z$ is closer to $1/n$ compared to $y$, and estimated statistics $z_{approx}$ is also closer to $1/n$ than $y_{approx}$.

In Figures \ref{toy2} and \ref{toy3}, we assume that in the mini-dataset, 5 samples are sampled at each step, simulating the process where the agent randomly walks and visits all states in the environment, we initialized target networks for DRND, RDD to $f_\text{tar}(s) \sim \mathcal{N}(\mu\times\bm{1_d},\sigma^2\times\bm{1_d})$. In Figure \ref{toy2}, we set $\mu=1, \sigma=0.8, \bm{d}=256,$ and set the numbers of target networks $N=100$ for DRND. In Figure \ref{toy3}, we initialized $\mu=1, \sigma=0.4, \bm{d}=256,$ and set the numbers of target networks $N=200$ for DRND. From these two examples, when using the estimation method that calculates statistics through a prediction network, RDD has better tracking ability compared to DRND, and is closer to the $1/n$ of the virtual counting method. When the variance decreases, the bonus given by RDD for the state gradually approaches that of RND. For DRND, as the number of target networks increases, the estimated value of the bonus and the ground truth value are closer to RDD, thereby supporting the conclusion we presented in Section \ref{Relationship}. 

\begin{table}
\caption{The average episodic returns and standard errors of all models tested over 100 episodes.}
\label{tabel}
\centering
\scalebox{0.7}{ 
\begin{tabular}{ccccccc}
  \toprule
  Tasks & RDD & SASR \cite{ma2024highly} & DRND-online \cite{yang2024exploration} & ReLara \cite{ma2024reward} & GFA-RFE \cite{zhang2024uncertainty} & ROSA \cite{mguni2023learning} \\
  \hline
  Montezuma & \pmb{$8238.5\pm223.0$} & $6737.9\pm0.0$ & $6828.5\pm0.0$ & $2421.9\pm0.0$ & $4755.3\pm0.0$ & $4294.4\pm0.0$  \\
  PrivateEye & \pmb{$11223.2\pm1214.0$} & $7445.3\pm448.0$ & $6056.7\pm148.0$ & $4746\pm339.0$ & $2219\pm0.0$ & $1988\pm0.0$ \\
  Breakout & \pmb{$108.8\pm8.0$} & \pmb{$104.8\pm6.0$} & $69.4\pm5.0$ & $91.8\pm7.0$ & $56.4\pm2.0$ & $67.8\pm5.0$  \\
  Gravitar & $3342.1\pm214.0$ & $3041.1\pm123.0$ & \pmb{$3875.2\pm199.0$} & $2985.8\pm175.0$ & $1245.6\pm75.0$ & $985.6\pm66.0$  \\
  Adroit Pen & \pmb{$4010.5\pm34.66$} & $3099.3\pm45.22$ & $3891.5\pm23.67$ & $3465.7\pm24.38$ & $1924.2\pm87.99$ & $827.4\pm13.33$  \\
  Adroit Door & \pmb{$1934.1\pm21.34$} & $1867.3\pm24.12$ & \pmb{$1921.1\pm13.22$} & $1298.6\pm56.33$ & $998.7\pm13.67$ & $676.5\pm34.56$  \\
  Adroit Hammer & \pmb{$13221.2\pm123.22$} & $8899.2\pm123.12$ & $10234.1\pm133.89$ & $7892.8\pm98.22$ & $4321.9\pm78.33$ & $3368.1\pm56.34$  \\
  Adroit Relocate  & \pmb{$63.9\pm31.8$} & $43.1\pm25.2$ & $54.3\pm22.7$ & $47.4\pm31.7$ & $7.5\pm1.7$ & $3.5\pm1.1$  \\
  Fetch Reach & \pmb{$197.98\pm9.9$} & $81.29\pm6.52$ & $99.8\pm0.0$ & $187.9\pm0.0$ & $42.1\pm0.0$ & $0.1\pm0.0$ \\
  Fetch Push & \pmb{$191.7\pm4.5$} & $137.06\pm12.66$ & $122.2\pm0.0$ & $166.9\pm0.0$ & $49.1\pm0.0$ & $0.0\pm0.0$  \\
  Fetch Slide & \pmb{$147.9\pm12.0$} & $132.3\pm1.3$ & $127.2\pm0.0$ & $111.6\pm2.0$ & $115.8\pm2.0$ & $11.2\pm0.9$  \\
  Fetch PickPlace & \pmb{$1.0\pm0.0$} & \pmb{$1.0\pm0.0$} & \pmb{$1.0\pm0.0$} & \pmb{$1.0\pm0.0$} & $0.5\pm0.0$ & $0.0\pm0.0$  \\
  \bottomrule
   & ExploRS \cite{devidze2022exploration} & $\#$Explo \cite{tang2017exploration} & RND \cite{burda2018exploration} & SAC \cite{haarnoja2018soft} & TD3 \cite{fujimoto2018addressing} & PPO \cite{schulman2017proximal} \\
   \hline
   Montezuma & $3491.5\pm0.0$ & $1400.1\pm0.0$ & $5494.3\pm0.0$ & $0.0\pm0.0$ & $0.0\pm0.0$ & $0.0\pm0.0$ \\
   PrivateEye & $987\pm23.0$ & $879\pm18.0$ & $3318\pm125.0$ & $276\pm0.0$ & $314\pm0.0$ & $198\pm0.0$ \\
   Breakout & $45.5\pm8.0$ & $22.8\pm3.0$ & $53.8\pm3.0$ & $45.6\pm6.0$ & $39.2\pm3.0$ & $78.8\pm7.0$ \\
   Gravitar & $675.8\pm45.0$ & $566.8\pm34.0$ & $3129.8\pm185.0$ & $295.8\pm15.0$ & $123.8\pm17.0$ & $98.8\pm12.0$ \\
   Adroit Pen & $555.6\pm23.33$ & $325.9\pm24.18$ & $1925.2\pm34.22$ & $125.2\pm13.92$ & $34.56\pm3.42$ & $154.24\pm5.62$ \\
   Adroit Door & $567.2\pm4.55$ & $434.4\pm3.24$ & $123.2\pm0.00$ & $67.7\pm0.00$ & $58.6\pm0.00$ & $143.0\pm0.00$ \\
   Adroit Hammer & $2321.2\pm67.19$ & $1850.90\pm45.66$ & $6521.7\pm78.72$ & $1276.4\pm23.65$ & $897.4\pm19.34$ & $597.1\pm15.01$ \\
   Adroit Relocate  & $3.3\pm0.3$ & $3.1\pm0.0$ & $23.8\pm4.9$ & $4.9\pm0.7$ & $12.3\pm0.2$ & $13.4\pm0.5$ \\
   Fetch Reach & $0.7\pm0.0$ & $4.6\pm0.0$ & $69.3\pm0.0$ & $156.5\pm0.0$ & $0.0\pm0.0$ & $79.5\pm0.0$ \\
   Fetch Push & $0.0\pm0.0$ & $3.7\pm0.0$ & $0.0\pm0.0$ & $0.0\pm0.0$ & $0.0\pm0.0$ & $0.0\pm0.0$ \\
   Fetch Slide & $4.3\pm0.1$ & $3.5\pm0.0$ & $4.8\pm0.2$ & $0.7\pm0.2$ & $0.5\pm0.4$ & $0.2\pm0.2$ \\
   Fetch PickPlace & $0.0\pm0.0$ & $0.0\pm0.0$ & $0.0\pm0.0$ & $0.0\pm0.0$ & $0.0\pm0.0$ & $0.0\pm0.0$ \\
   \bottomrule
\end{tabular}
}
\end{table}

\subsection{Performance of Experiments}\label{Performance of Experiments}

We evaluate RDD methods across three challenging domains: (1) four Atari 2600 games including the notoriously difficult Montezuma's Revenge \cite{bellemare2013arcade}, (2) four Adroit robotic manipulation tasks \cite{rajeswaran2017learning}, and (3) four Fetch robotic manipulation tasks \cite{plappert2018multi}. All environments feature sparse reward structures: most provide only a +1 reward for successful task completion (except Breakout), while Adroit tasks offer +10.0 for success and -0.1 otherwise. To ensure statistical reliability, we perform five independent runs per task with different random seeds and report mean performance. We maintain consistent hyperparameter settings and neural network architectures across all evaluated tasks, with specific details of environments tasks and hyperparameters provided in the Appendix \ref{details} and \ref{hyper}.

\begin{figure}[htp]
  \centering
  \begin{subfigure}{0.23\textwidth}
    \centering
    \includegraphics[width=1\linewidth]{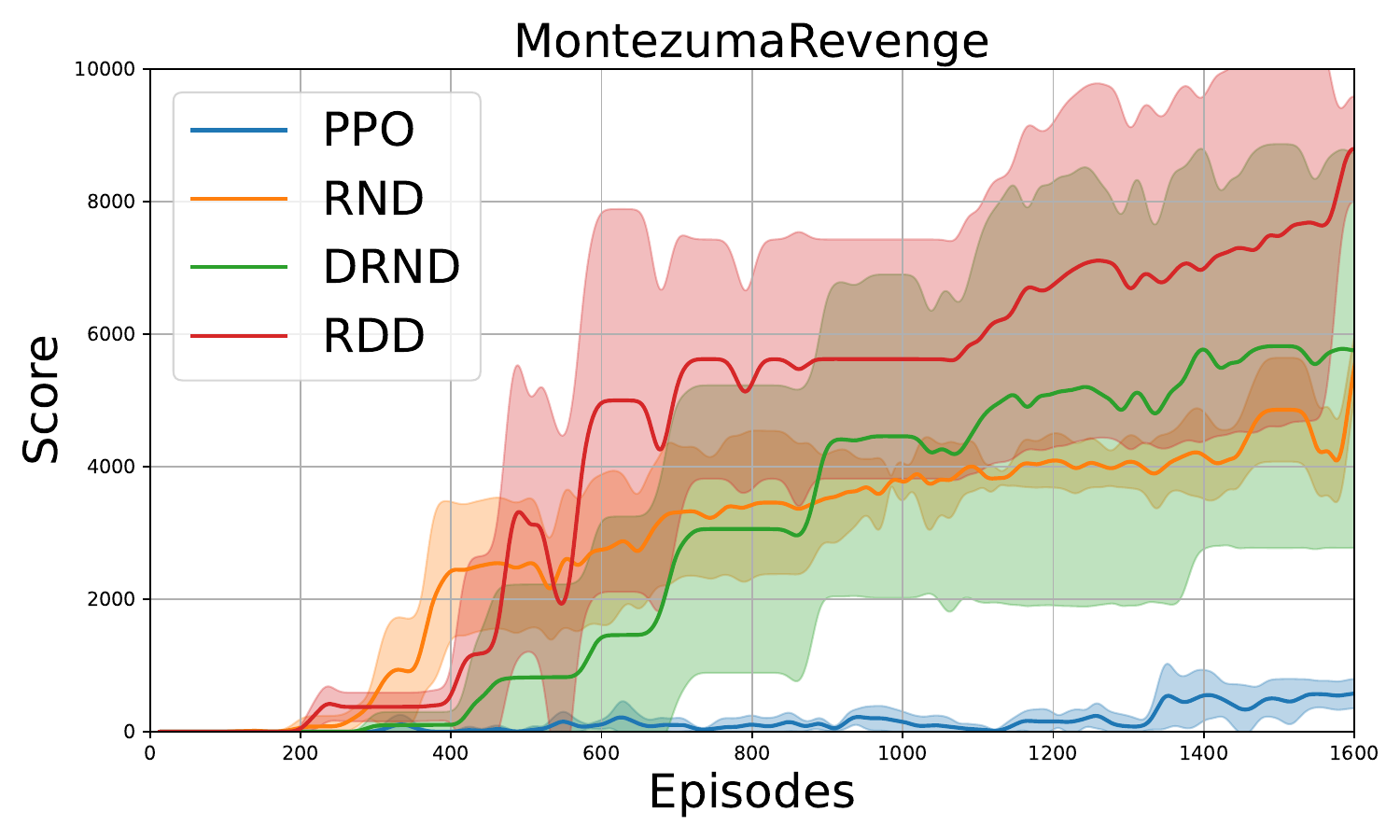}
    \captionsetup{font=scriptsize}
  \end{subfigure}
  \begin{subfigure}{0.23\textwidth}
    \centering
    \includegraphics[width=1\linewidth]{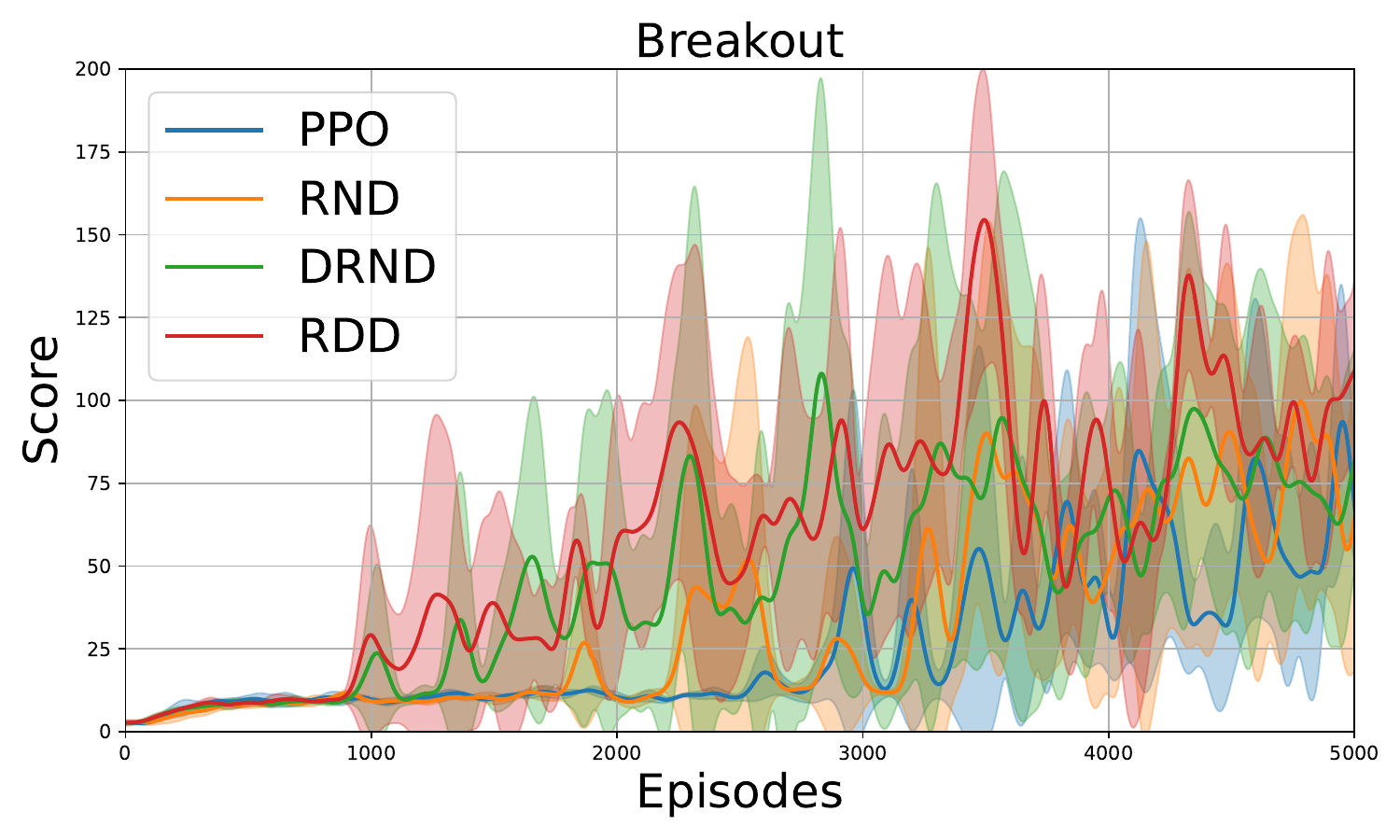}
    \captionsetup{font=scriptsize}
  \end{subfigure}
  \begin{subfigure}{0.23\textwidth}
    \centering
    \includegraphics[width=1\linewidth]{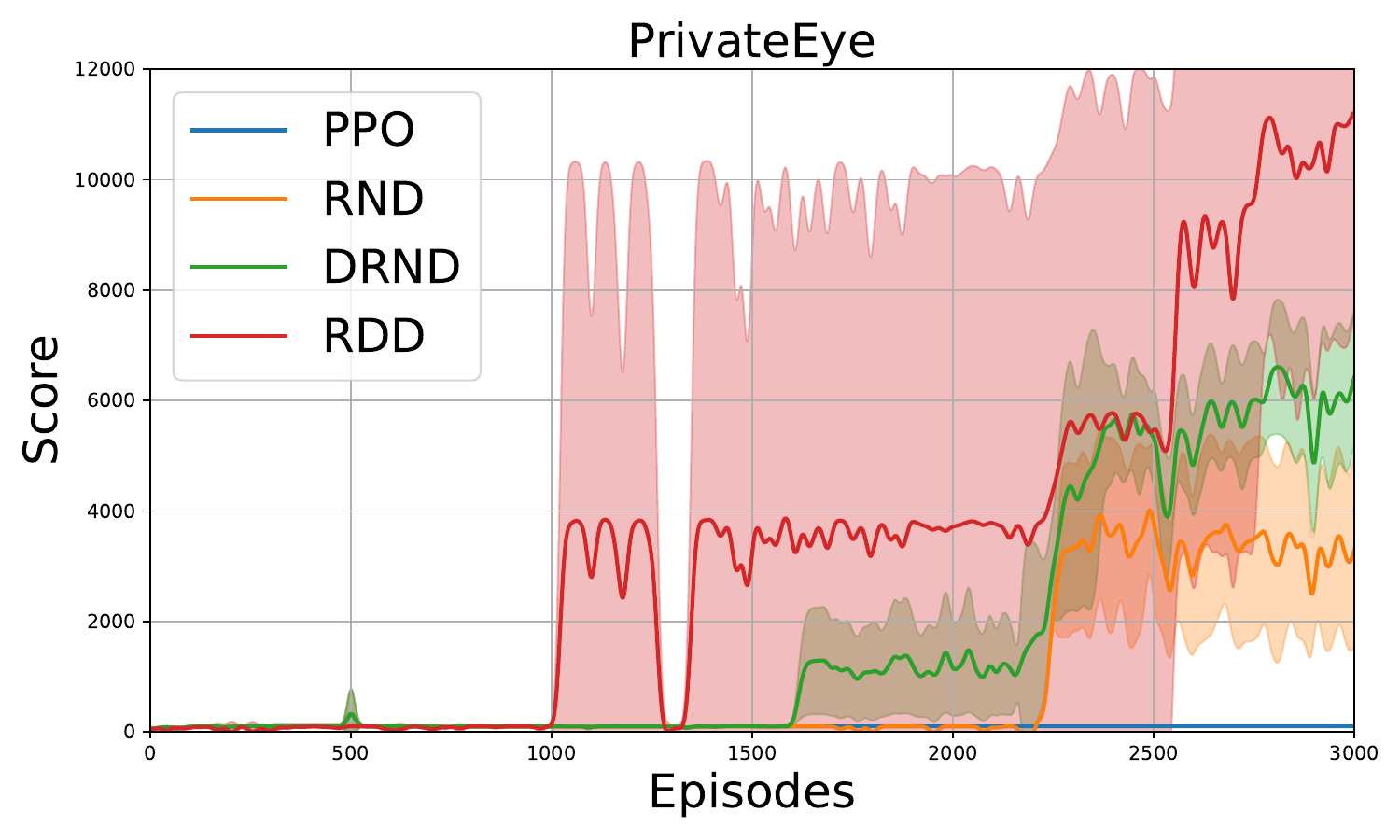} 
    \captionsetup{font=scriptsize}
  \end{subfigure}
  \begin{subfigure}{0.23\textwidth}
    \centering
    \includegraphics[width=1\linewidth]{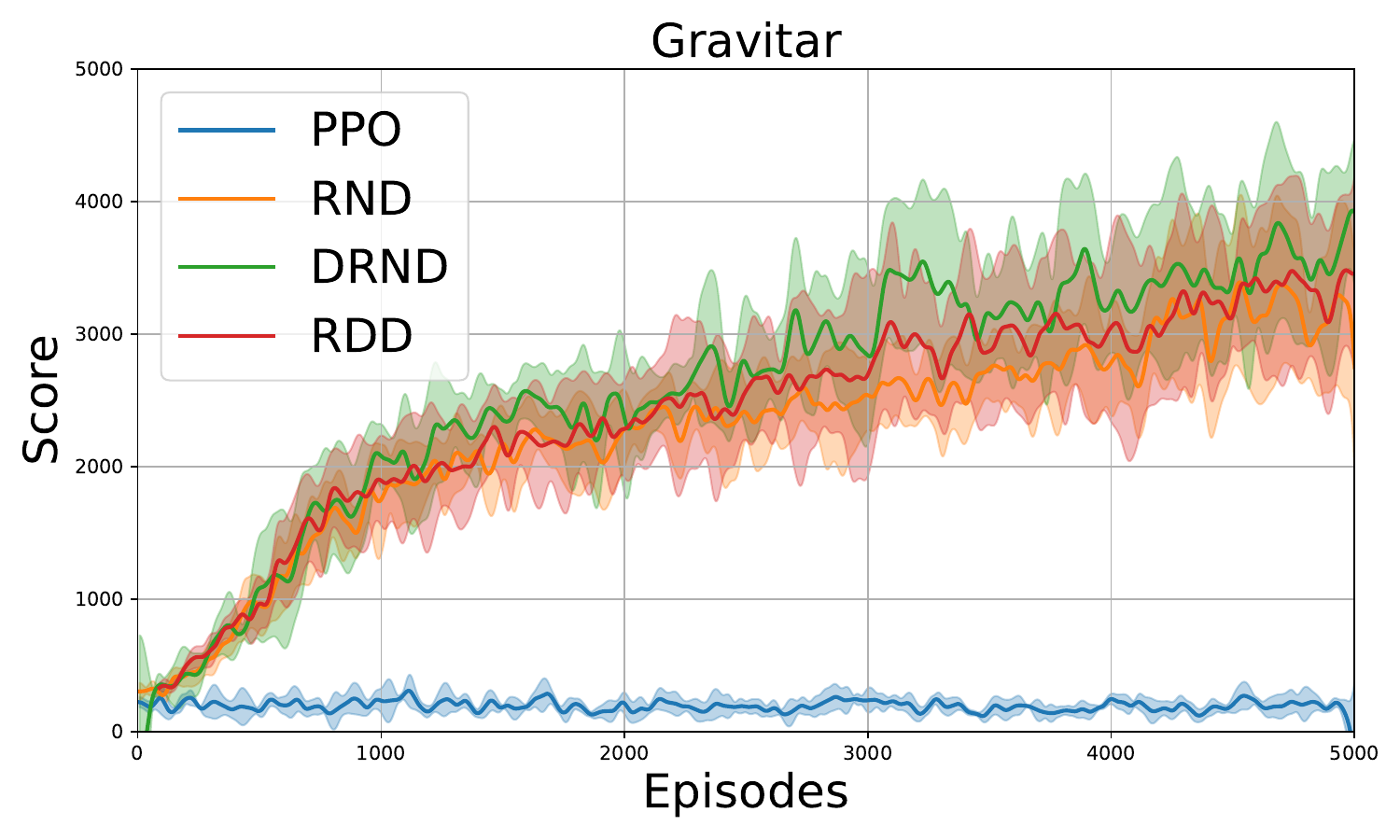} 
    \captionsetup{font=scriptsize}
  \end{subfigure}\\
\vspace{-0.1cm}
  \begin{subfigure}{0.23\textwidth}
    \centering
    \includegraphics[width=1\linewidth]{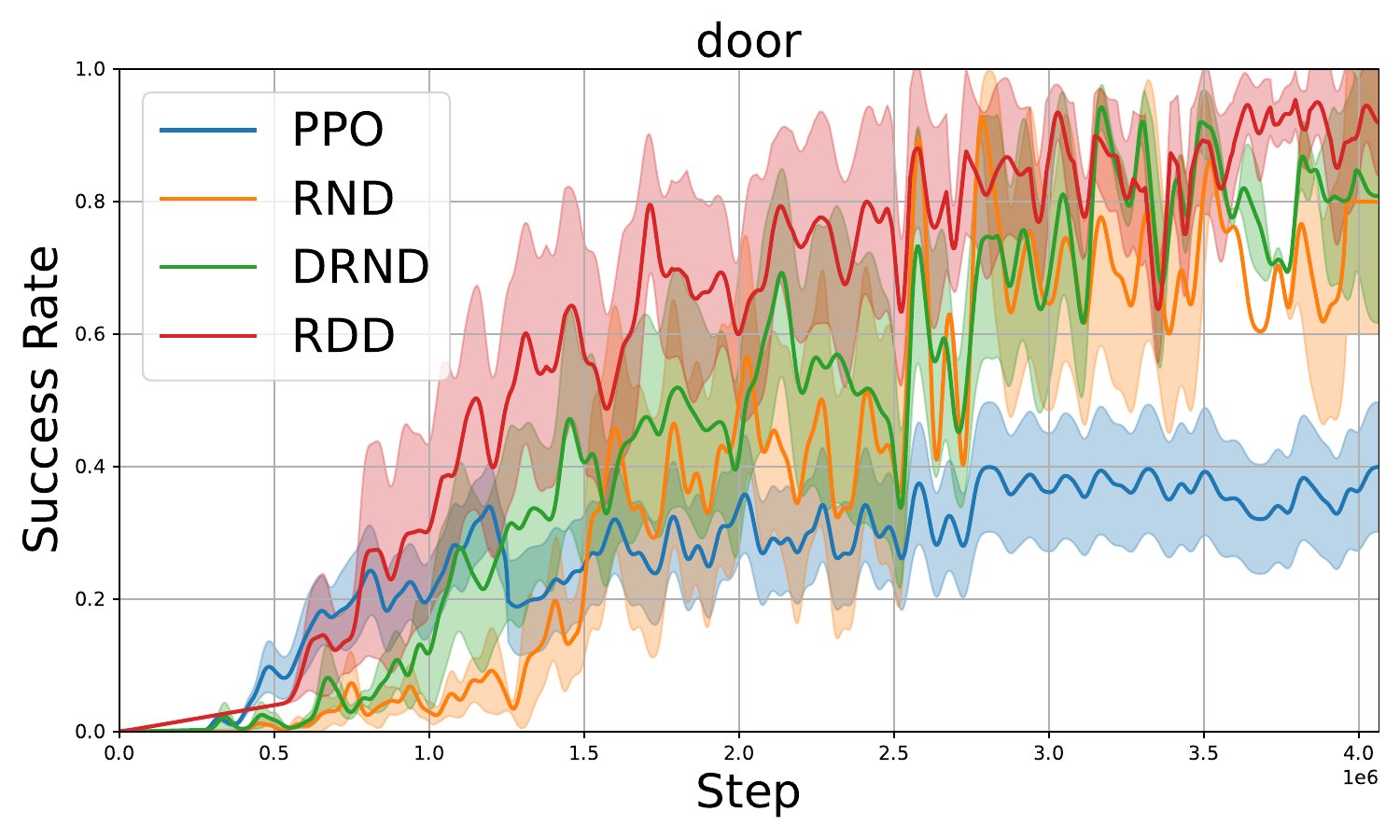}
    \captionsetup{font=scriptsize}
  \end{subfigure}
  \begin{subfigure}{0.23\textwidth}
    \centering
    \includegraphics[width=1\linewidth]{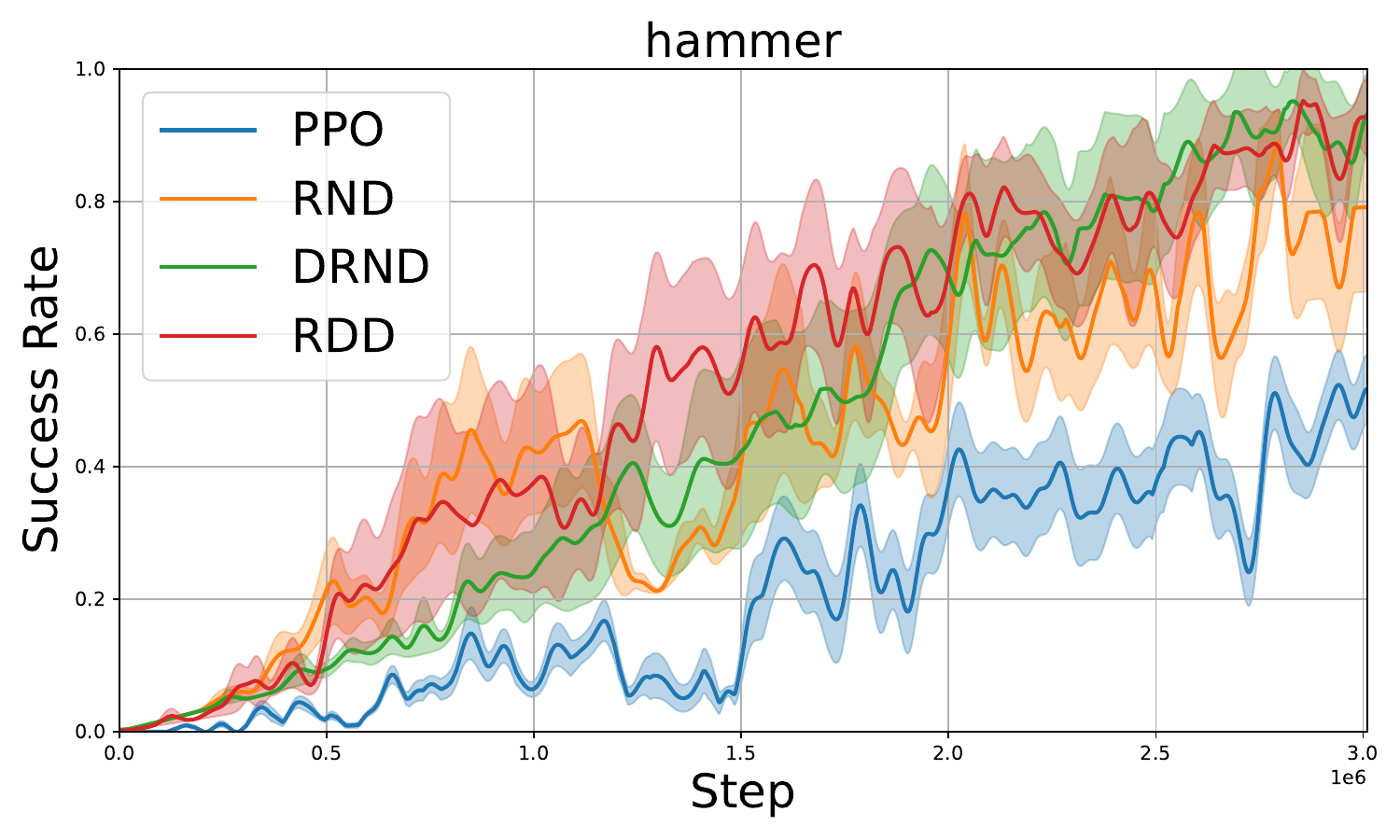} 
    \captionsetup{font=scriptsize}
  \end{subfigure}
  \begin{subfigure}{0.23\textwidth}
    \centering
    \includegraphics[width=1\linewidth]{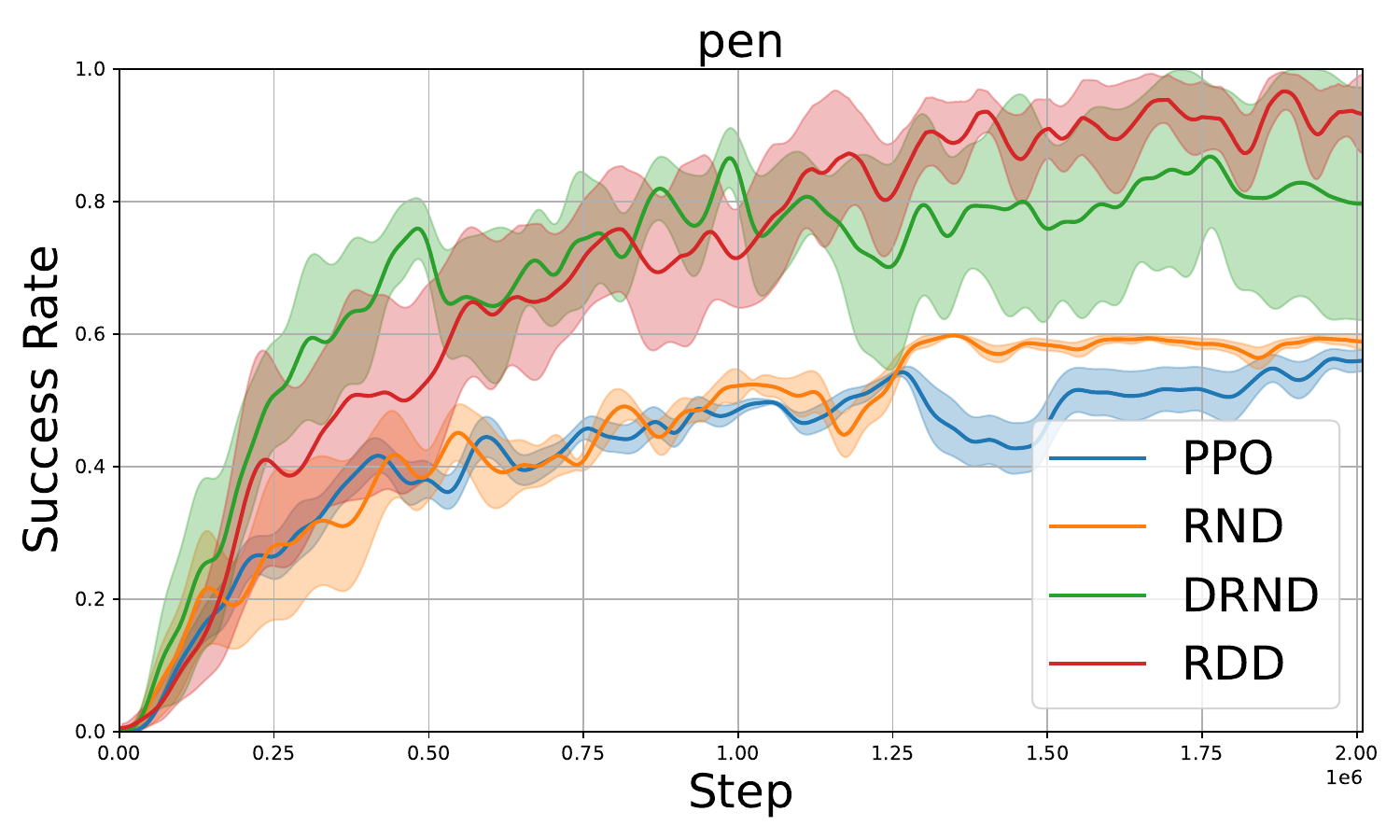} 
    \captionsetup{font=scriptsize}
  \end{subfigure}
  \begin{subfigure}{0.23\textwidth}
    \centering
    \includegraphics[width=1\linewidth]{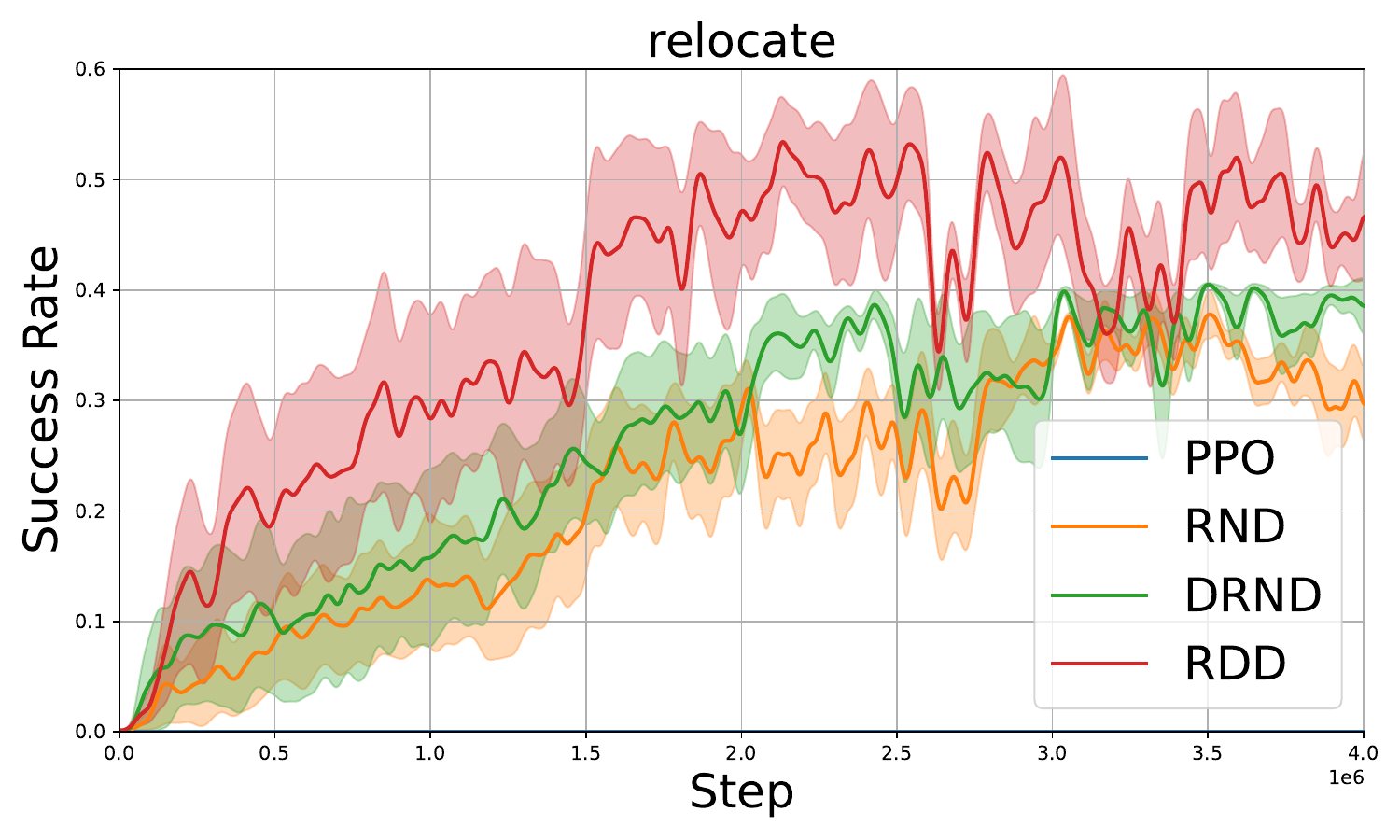} 
    \captionsetup{font=scriptsize}
  \end{subfigure}\\
\vspace{-0.1cm}
  \begin{subfigure}{0.23\textwidth}
    \centering
    \includegraphics[width=1\linewidth]{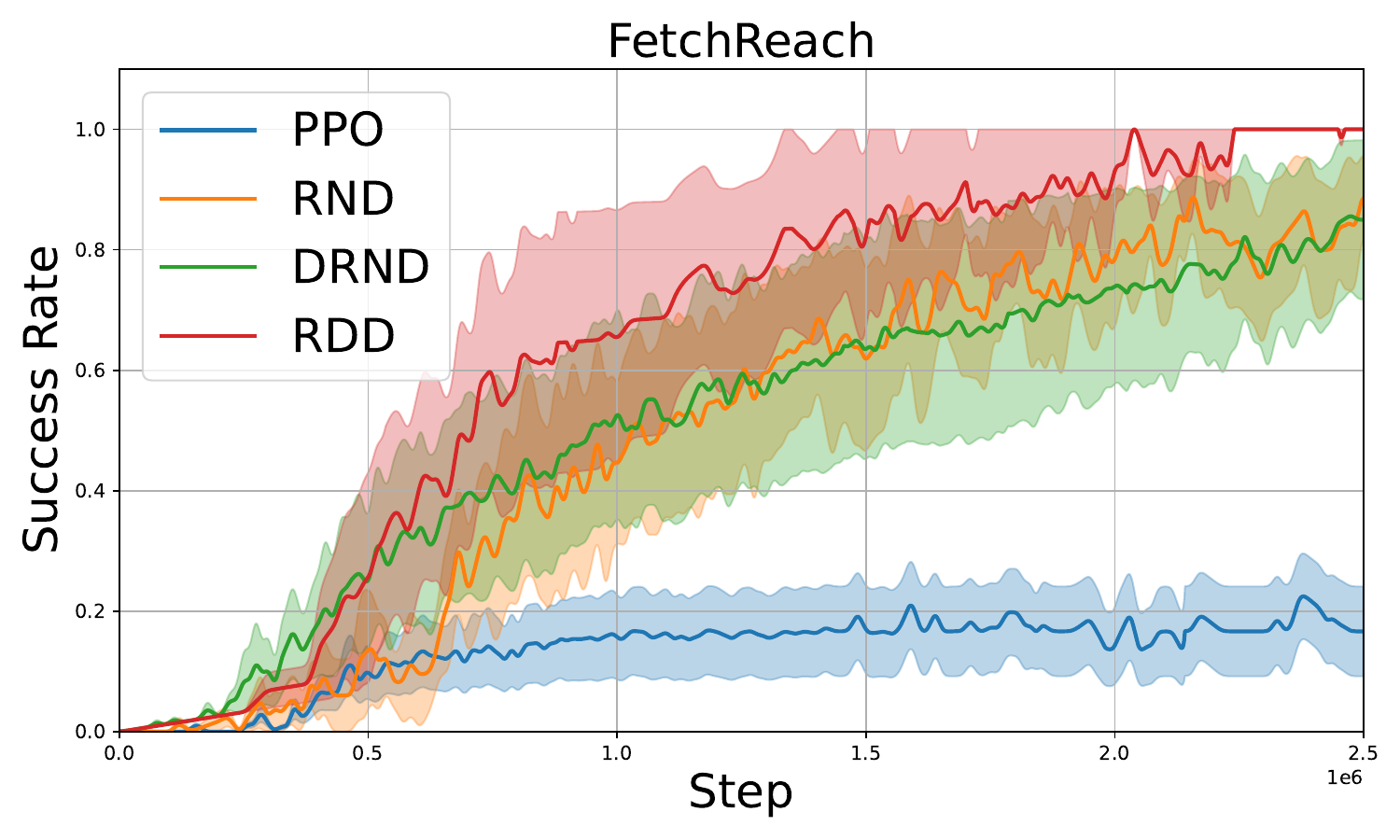}
    \captionsetup{font=scriptsize}
  \end{subfigure}
  \begin{subfigure}{0.23\textwidth}
    \centering
    \includegraphics[width=1\linewidth]{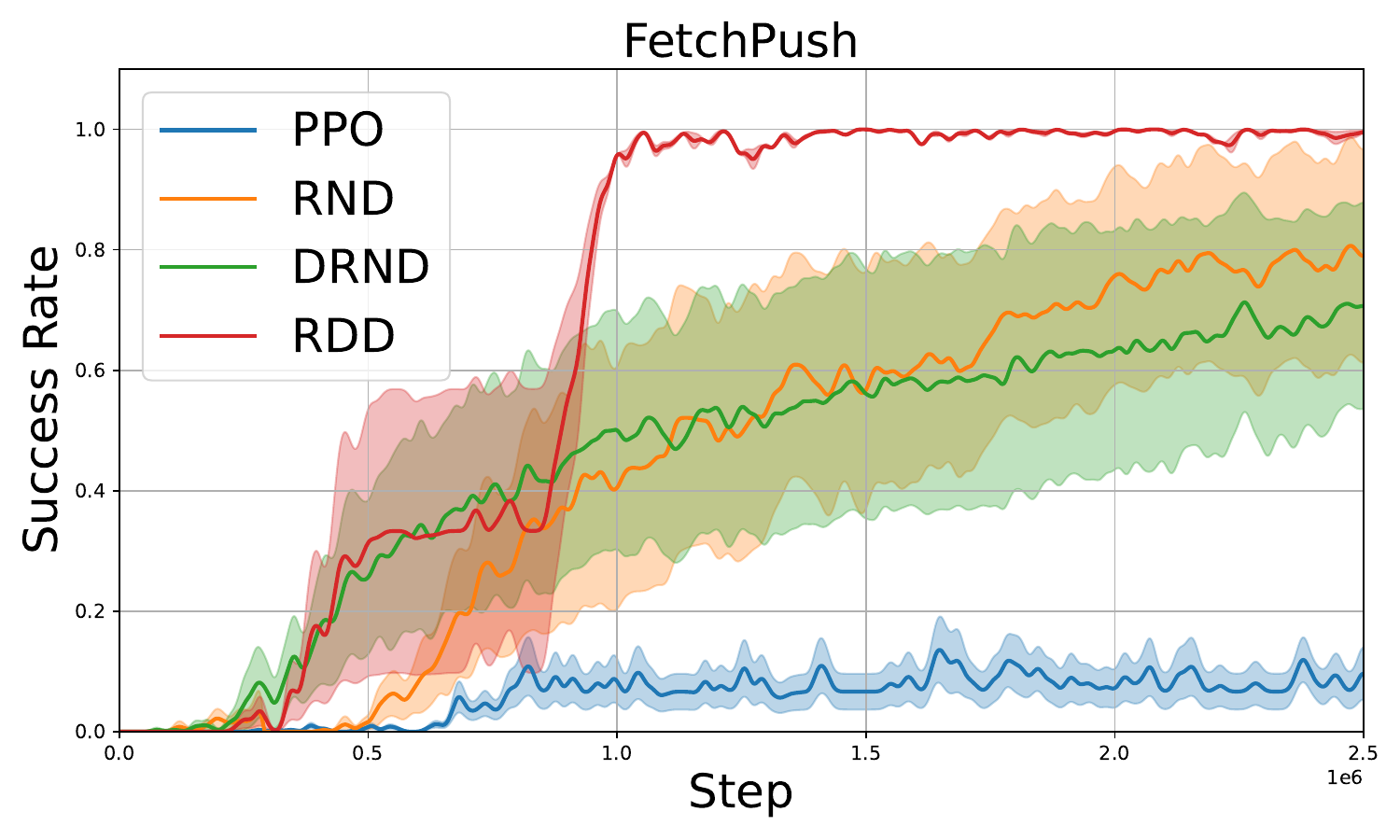} 
    \captionsetup{font=scriptsize}
  \end{subfigure}
  \begin{subfigure}{0.23\textwidth}
    \centering
    \includegraphics[width=1\linewidth]{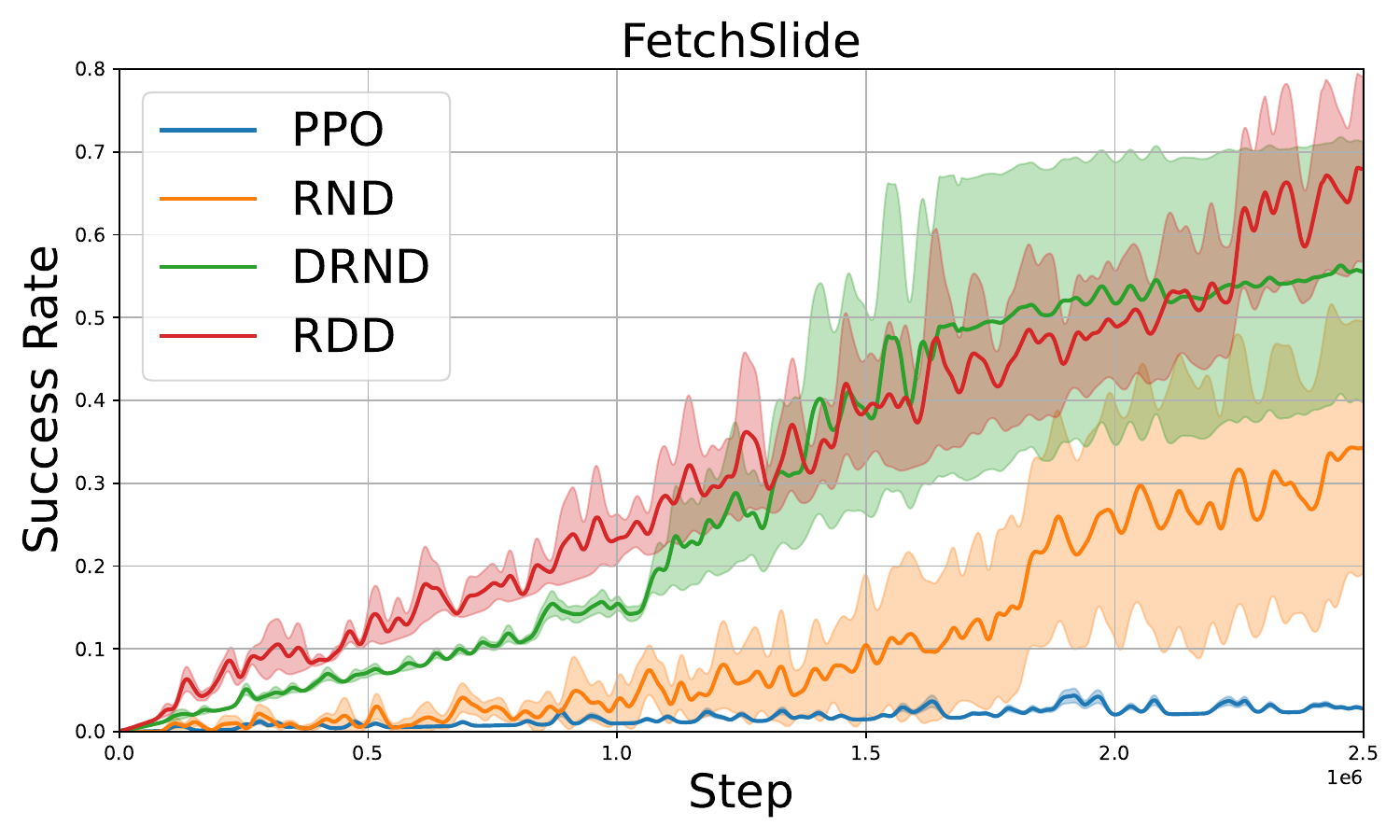} 
    \captionsetup{font=scriptsize}
  \end{subfigure}
  \begin{subfigure}{0.23\textwidth}
    \centering
    \includegraphics[width=1\linewidth]{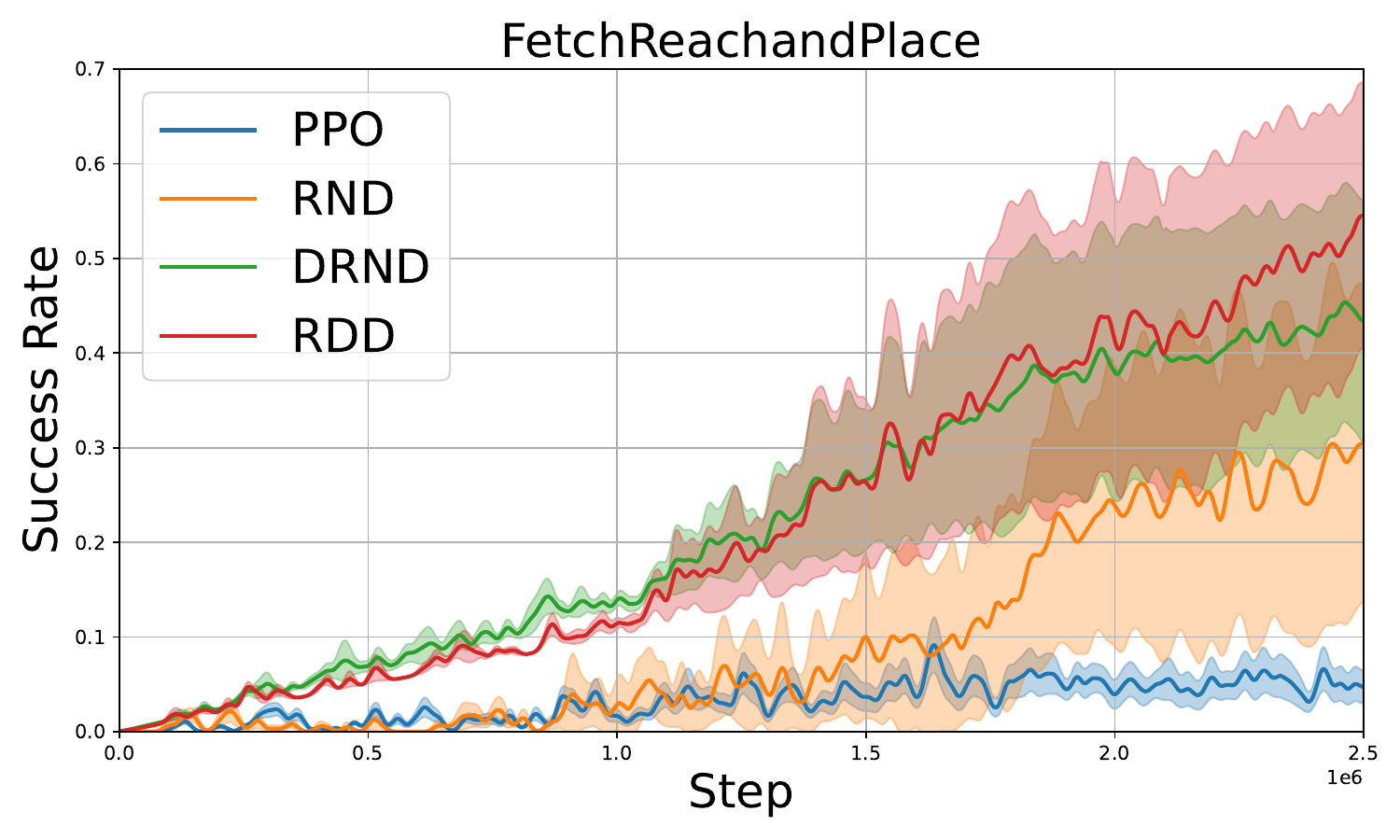} 
    \captionsetup{font=scriptsize}
  \end{subfigure}
\caption{Main experiment. We selected the currently most effective prediction-error method on 12 different environments for the experiment. Most of the  environments we used were extremely sparse reward.}
\label{main}
\vspace{-0.5cm}
\end{figure}

The results in Atari games are shown in the first line of Fig. \ref{main}. RDD demonstrated faster convergence and superior final scores compared to DRND, RND, and PPO. We further assessed RDD on Adroit continuous control tasks (second line of Fig. \ref{main}), where it outperformed other methods in exploration for 'Relocate', but showed comparable performance in the simpler 'Hammer' environment. Finally, the effectiveness of RDD in complex robotic manipulation was validated in Fetch tasks (third line of Fig. \ref{main}), where it excelled extremely in FetchPush environments, leading to significant improvements over DRND and other algorithms.


RDD demonstrated superior sampling efficiency, learning stability, and convergence speed compared to the baseline. This improvement can be attributed to RDD's constraint on the output distribution of the network, which effectively acts as a regularizer, unlike prior methods employing a target network. Furthermore, RDD utilizes a single target network, preventing the challenges associated with a single deterministic network attempting to fit the sampling of multiple target networks (thereby reducing uncertainty in the learning objective). Lastly, RDD entirely omits the difference between the target and predictor networks as a bonus component, which mitigates the issue of disproportionate bonus magnitudes across states with equivalent visit counts, which has been shown in Fig \ref{toy1}.



\subsection{Case Study}
\begin{figure}[htp]
\centering
\includegraphics[width=0.8\linewidth]{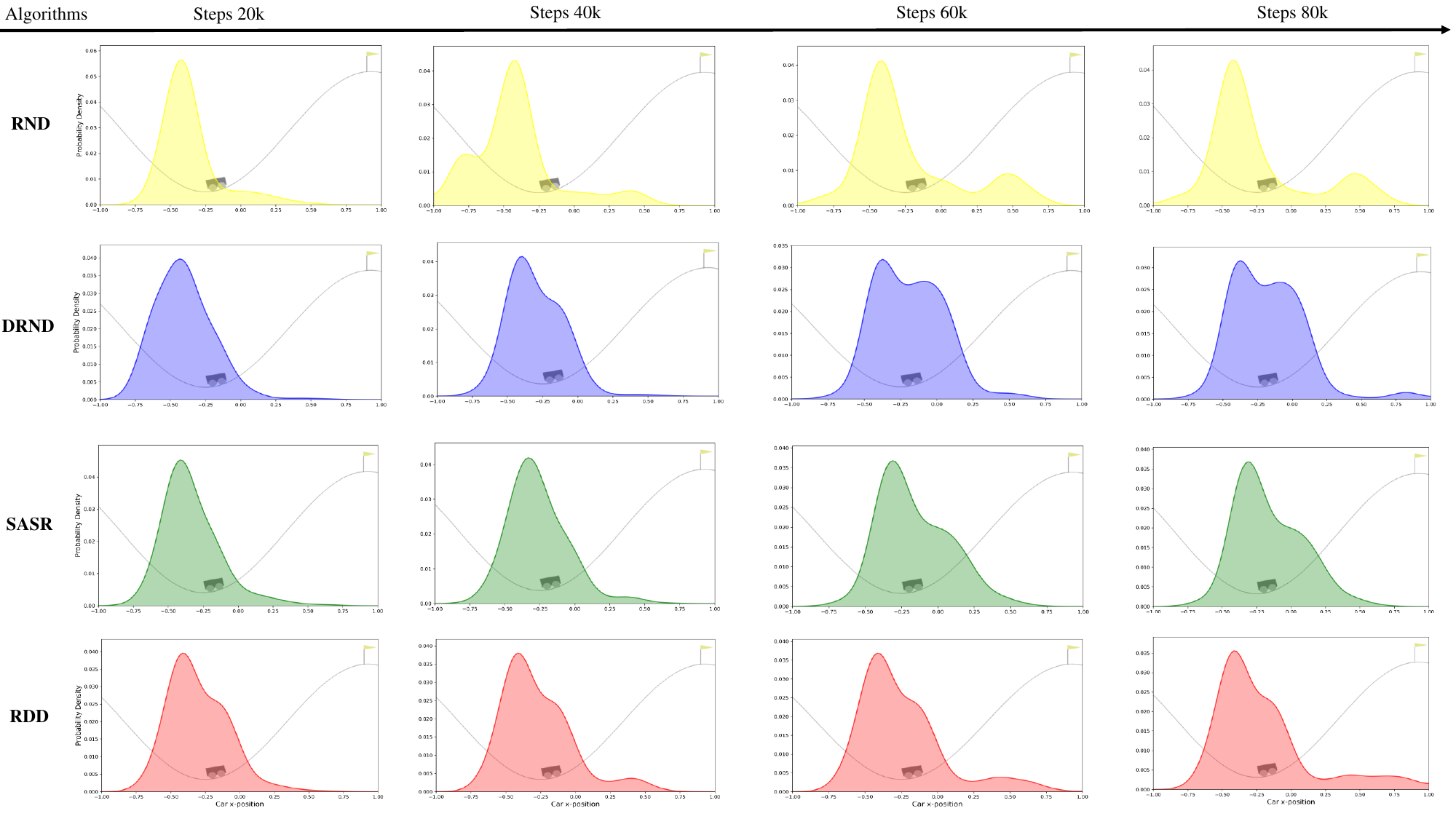} 
\caption{Visualize the exploration process on the MountainCar. In the MountainCar environment, the position of the car on the x-axis can reflect the success of the task. We represent the degree of exploration through the probability density of the agents of each method on the x-axis at different steps.}
\label{mountaincar}
\end{figure}

To further investigate RDD’s exploration behavior, we compare the visited state density throughout training in the MountainCar \cite{towers2024gymnasium} task with four representative exploration and RL strategies: (1) SASR \cite{ma2024highly}, use reward shaping for efficient self-adaptive reward; (2) DRND and (3) RND, which rewards novel states; The probability distribution of the x-coordinate of the car for every 20k steps is shown in Figure \ref{mountaincar}. We observe that RDD progressively covers a wider range of the state space. And in the later stage of the exploration, the probability density of the small car appearing around 1.0 (for completing the task) was much higher than that of other methods. In terms of the number of coverage states, DRND, SASR and RDD are roughly the same. However, compared with the other two methods, RDD has higher sample efficiency, as it can reach the target state at more positions in fewer steps. At the same time, we can see that RND and SASR have a narrower exploration range and are prone to getting stuck in local optimal solutions. Overall, RDD demonstrates a more effective exploration ability and can collect valuable samples more quickly, thereby achieving faster convergence.
\subsection{Ablation Study}
We conduct ablation studies to investigate key components of RDD. We select six representative tasks and report the experimental results. It is worth noting that due to the simplicity and effectiveness of the RDD, the number of hyper-parameters that have a significant impact on the algorithm's performance is also relatively small. Parameter tuning in the environment is also quite convenient.

\textbf{Variance of target network $\sigma$}
Under three experimental settings with variances of $\sigma=1$, $\sigma=0.1$, and a variable variance $\sigma=\sigma(s)$ (initializing a variance network), $\sigma=1$ achieved better results. We analyzed that during the exploration process, the variable variance might cause a fixed prediction network to predict a distribution, thereby leading to instability.

\textbf{Mean of target network $\mu$}
We examined the performance on six environments when the mean of the output of the target network was set to $0$, $1$, or $0.5$. It was found that the algorithm performed best when the mean was set to $1$. When the mean was $1$, for an unseen state, $z_n(s)$ would initially give a reward close to $1$ (achieved through $\|\bm{f}_{\theta}(s)- \bm{\mu}_{\bar\theta}(s)\|^2$), making the algorithm more sensitive to unseen states and showing a tendency to converge to a useful strategy more easily in the early stages of exploring the environment.

\textbf{Output dim of target network}
In terms of the output dimension of the target network, we adopted different hyperparameters depending on whether the output is image observations or state vectors for the environment. We found that when the output is image observations, with $dim=512$, and when it is the output state vector, with $dim=64$, better results were achieved.

\begin{figure}[htp]
  \centering
  \begin{subfigure}{0.15\textwidth}
    \centering
    \includegraphics[width=1.15\linewidth]{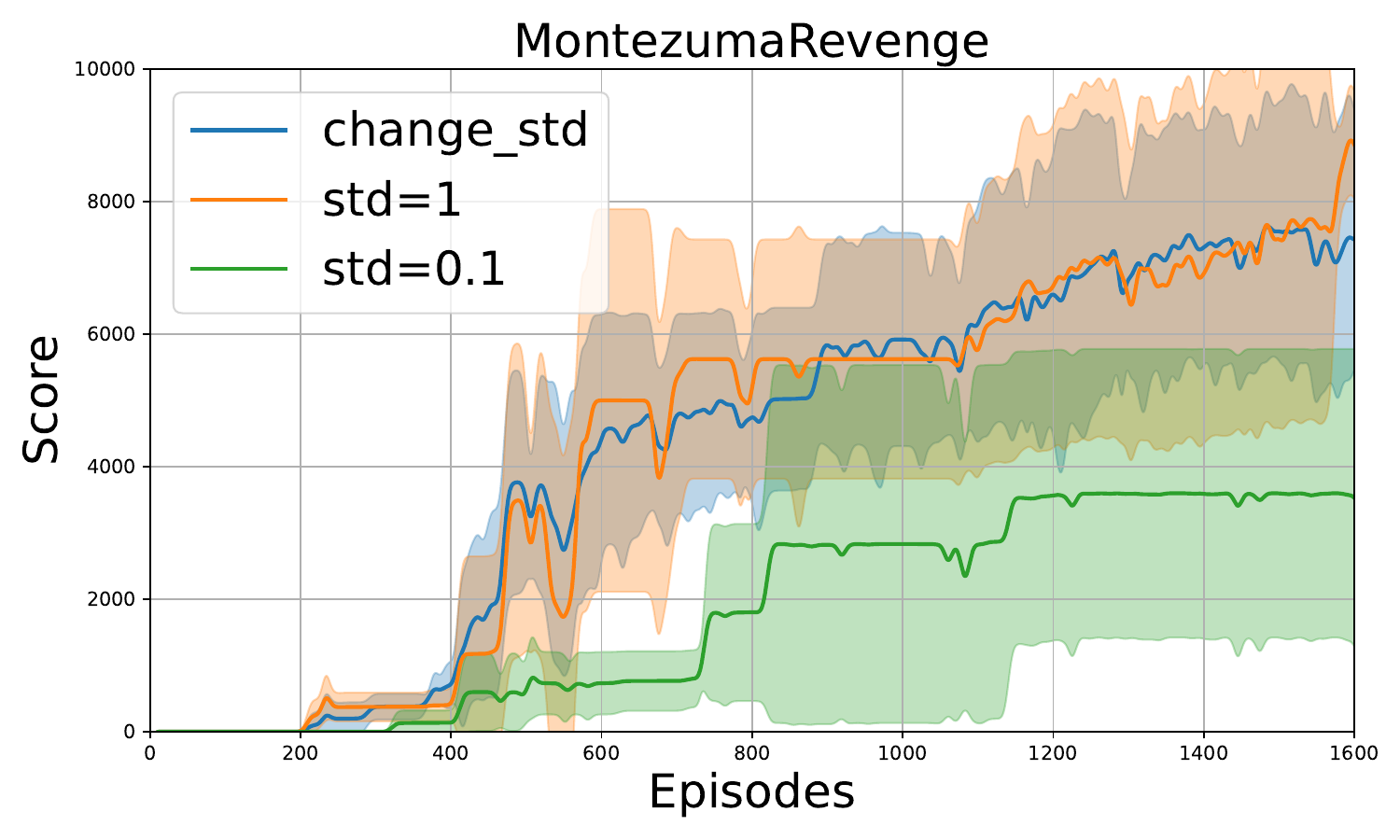}
    \captionsetup{font=scriptsize}
  \end{subfigure}
  \hspace{0.1cm}
  \begin{subfigure}{0.15\textwidth}
    \centering
    \includegraphics[width=1.15\linewidth]{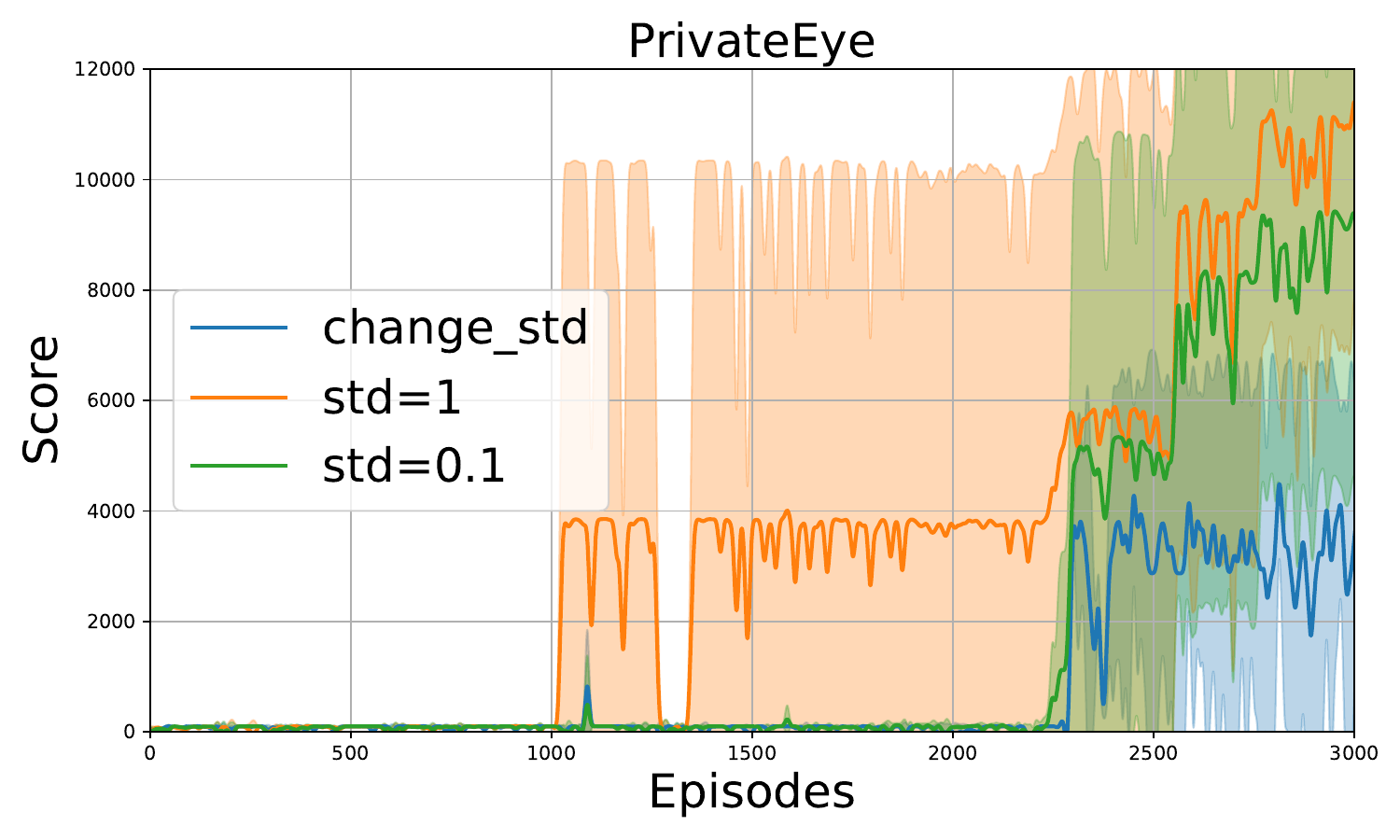}
    \captionsetup{font=scriptsize}
  \end{subfigure}
  \hspace{0.1cm}
  \begin{subfigure}{0.15\textwidth}
    \centering
    \includegraphics[width=1.15\linewidth]{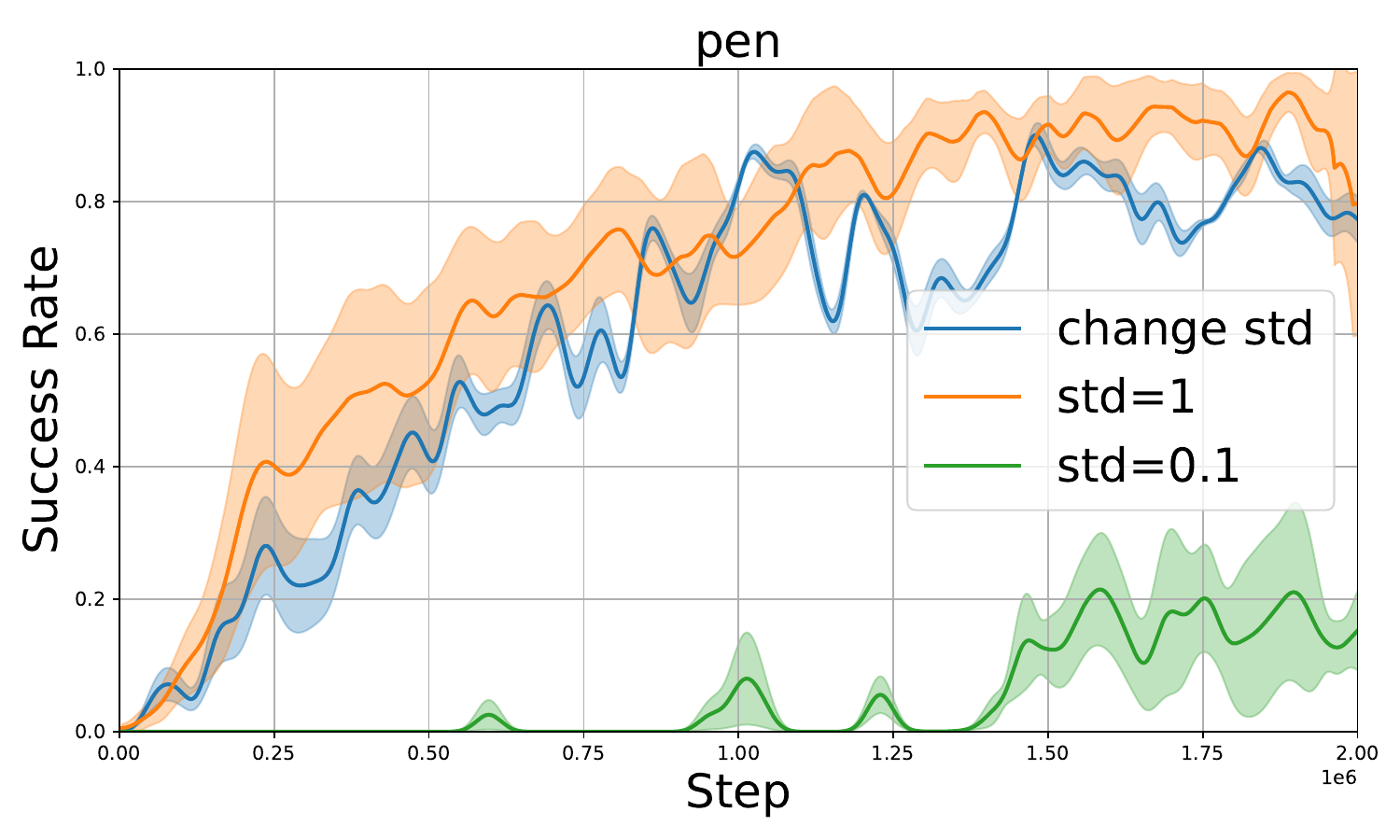} 
    \captionsetup{font=scriptsize}
  \end{subfigure}
  \hspace{0.1cm}
  \begin{subfigure}{0.15\textwidth}
    \centering
    \includegraphics[width=1.15\linewidth]{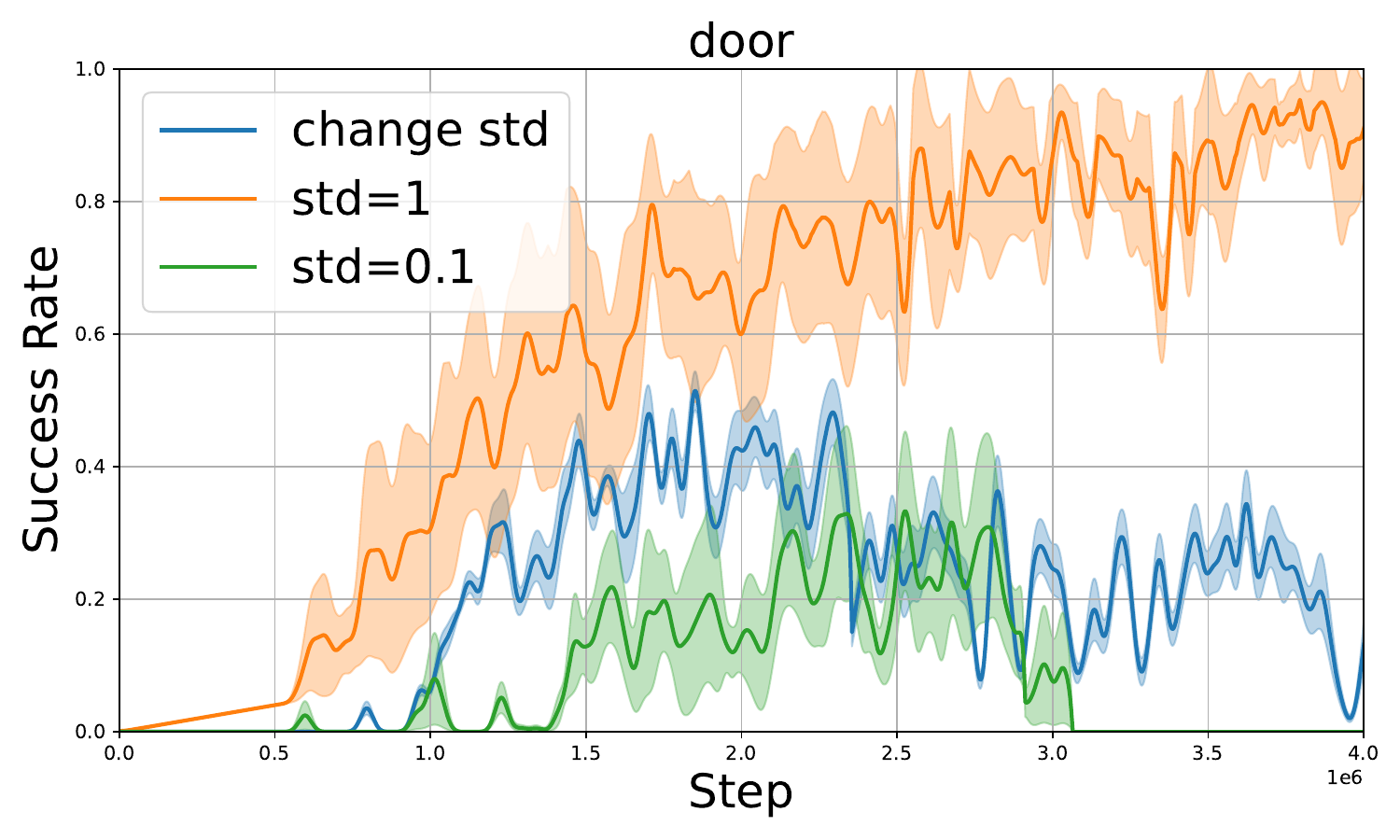} 
    \captionsetup{font=scriptsize}
  \end{subfigure}
  \hspace{0.1cm}
    \begin{subfigure}{0.15\textwidth}
    \centering
    \includegraphics[width=1.15\linewidth]{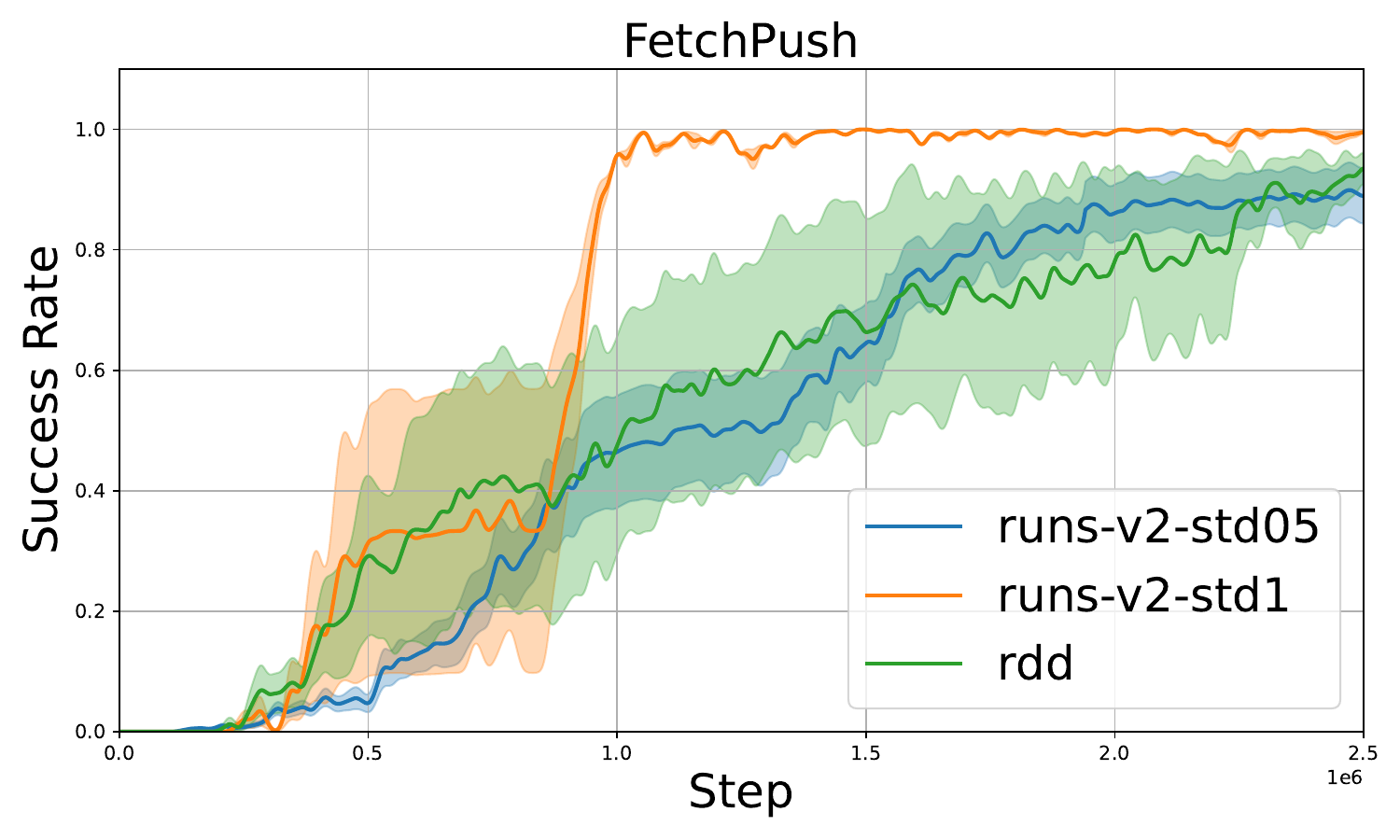} 
    \captionsetup{font=scriptsize}
  \end{subfigure}
  \hspace{0.1cm}
  \begin{subfigure}{0.15\textwidth}
    \centering
    \includegraphics[width=1.15\linewidth]{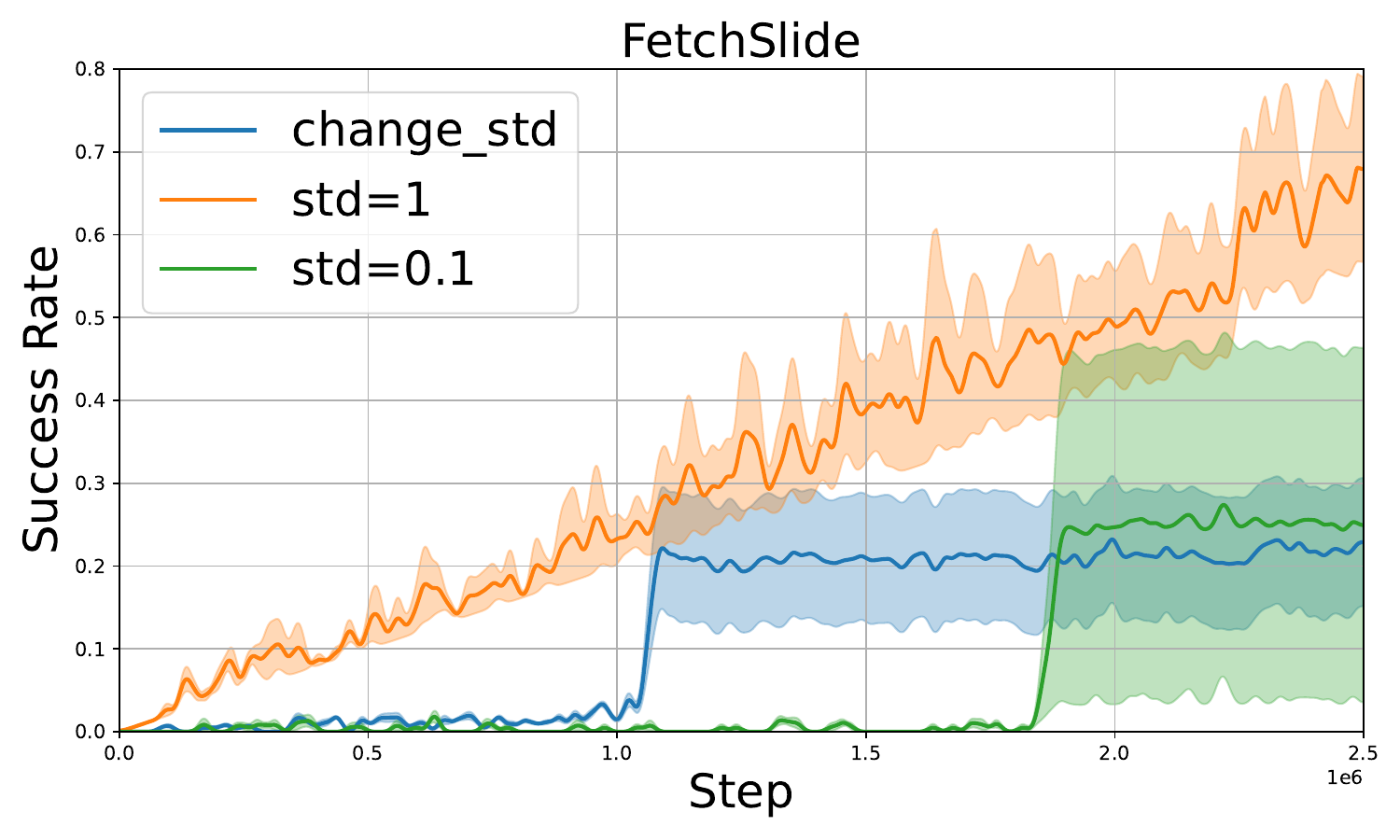} 
    \captionsetup{font=scriptsize}
  \end{subfigure}\\
  
      \begin{subfigure}{0.15\textwidth}
    \centering
    \includegraphics[width=1.15\linewidth]{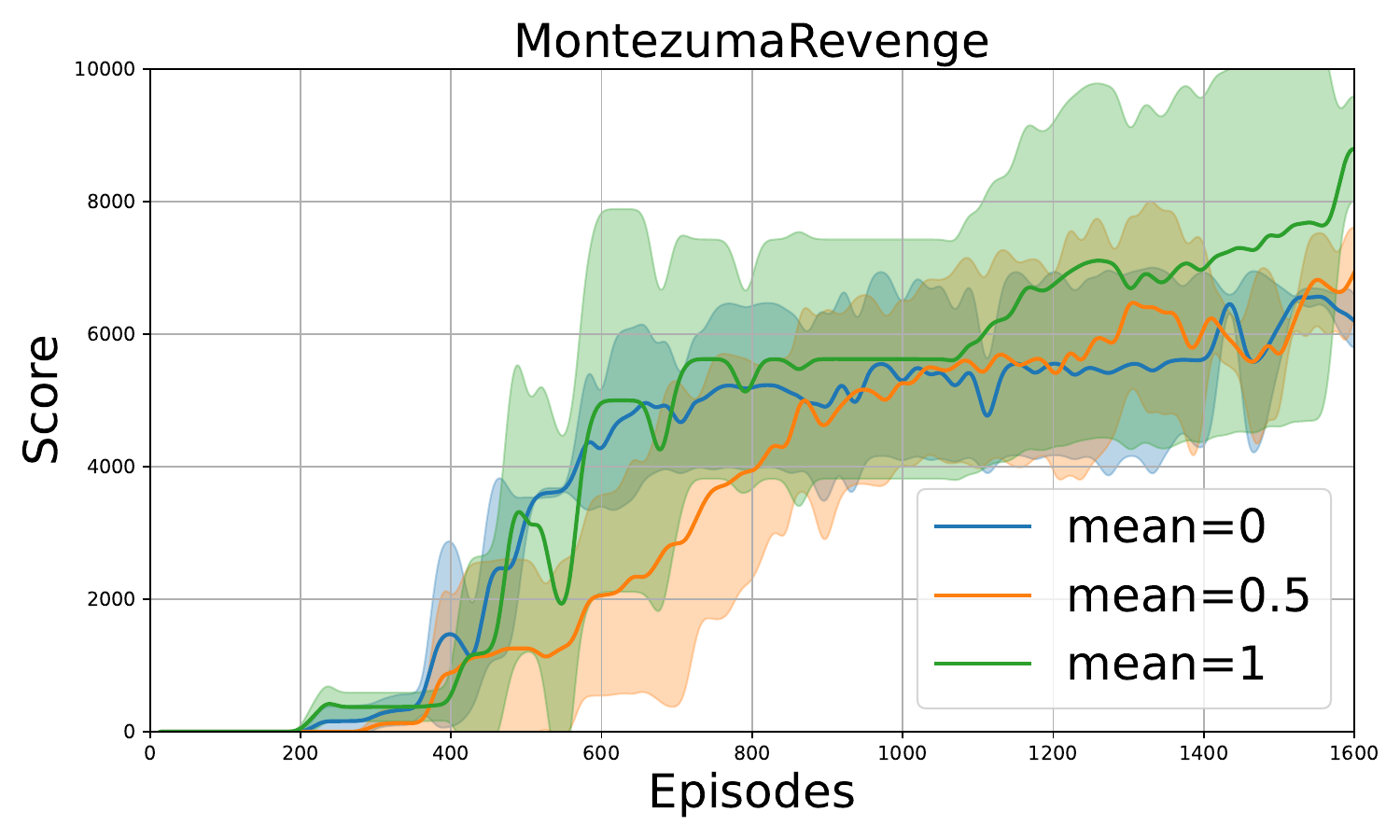}
    \captionsetup{font=scriptsize}
    
  \end{subfigure}
  \hspace{0.1cm}
  \begin{subfigure}{0.15\textwidth}
    \centering
    \includegraphics[width=1.15\linewidth]{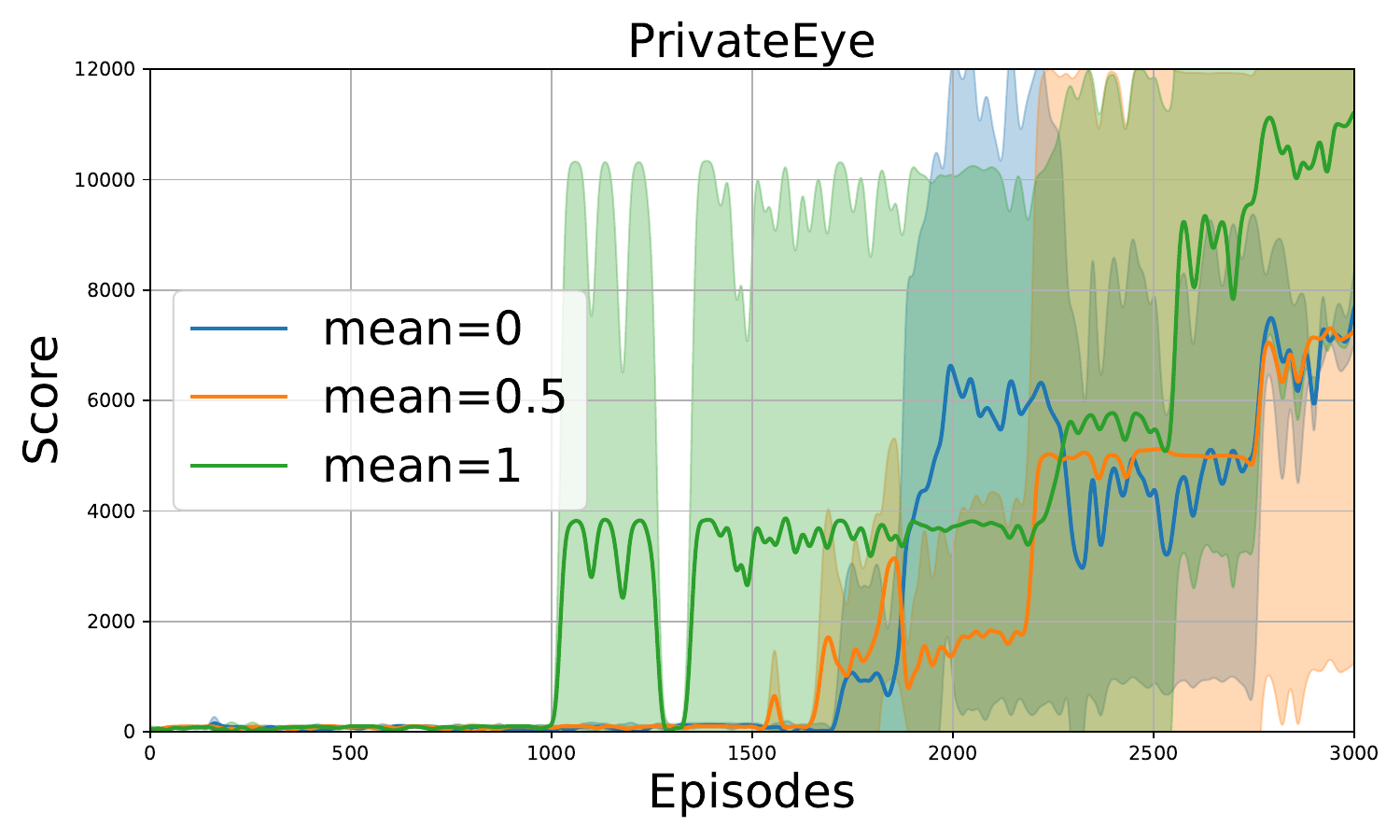}
    \captionsetup{font=scriptsize}
    
  \end{subfigure}
  \hspace{0.1cm}
  \begin{subfigure}{0.15\textwidth}
    \centering
    \includegraphics[width=1.15\linewidth]{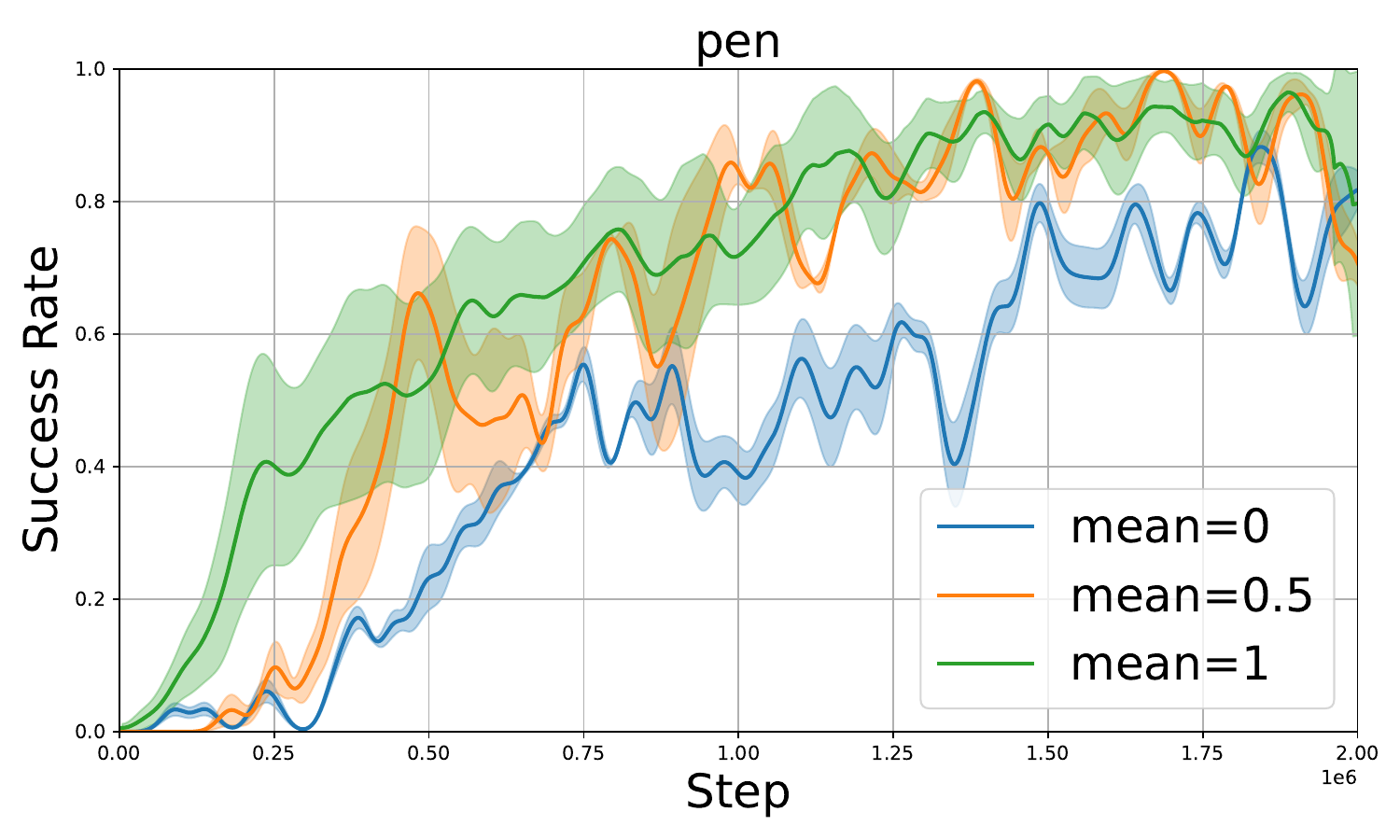} 
    \captionsetup{font=scriptsize}
    
  \end{subfigure}
  \hspace{0.1cm}
  \begin{subfigure}{0.15\textwidth}
    \centering
    \includegraphics[width=1.15\linewidth]{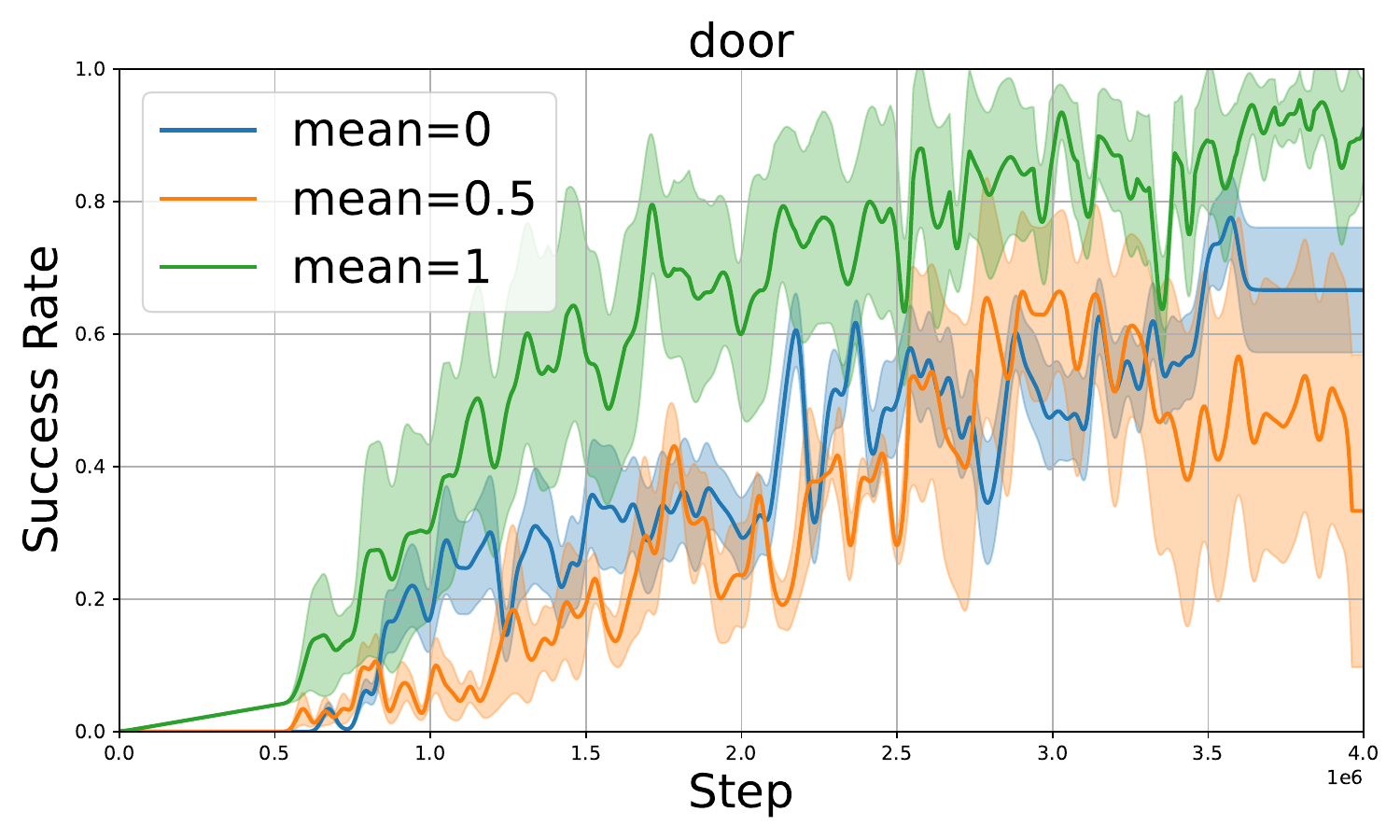} 
    \captionsetup{font=scriptsize}
    
  \end{subfigure}
  \hspace{0.1cm}
    \begin{subfigure}{0.15\textwidth}
    \centering
    \includegraphics[width=1.15\linewidth]{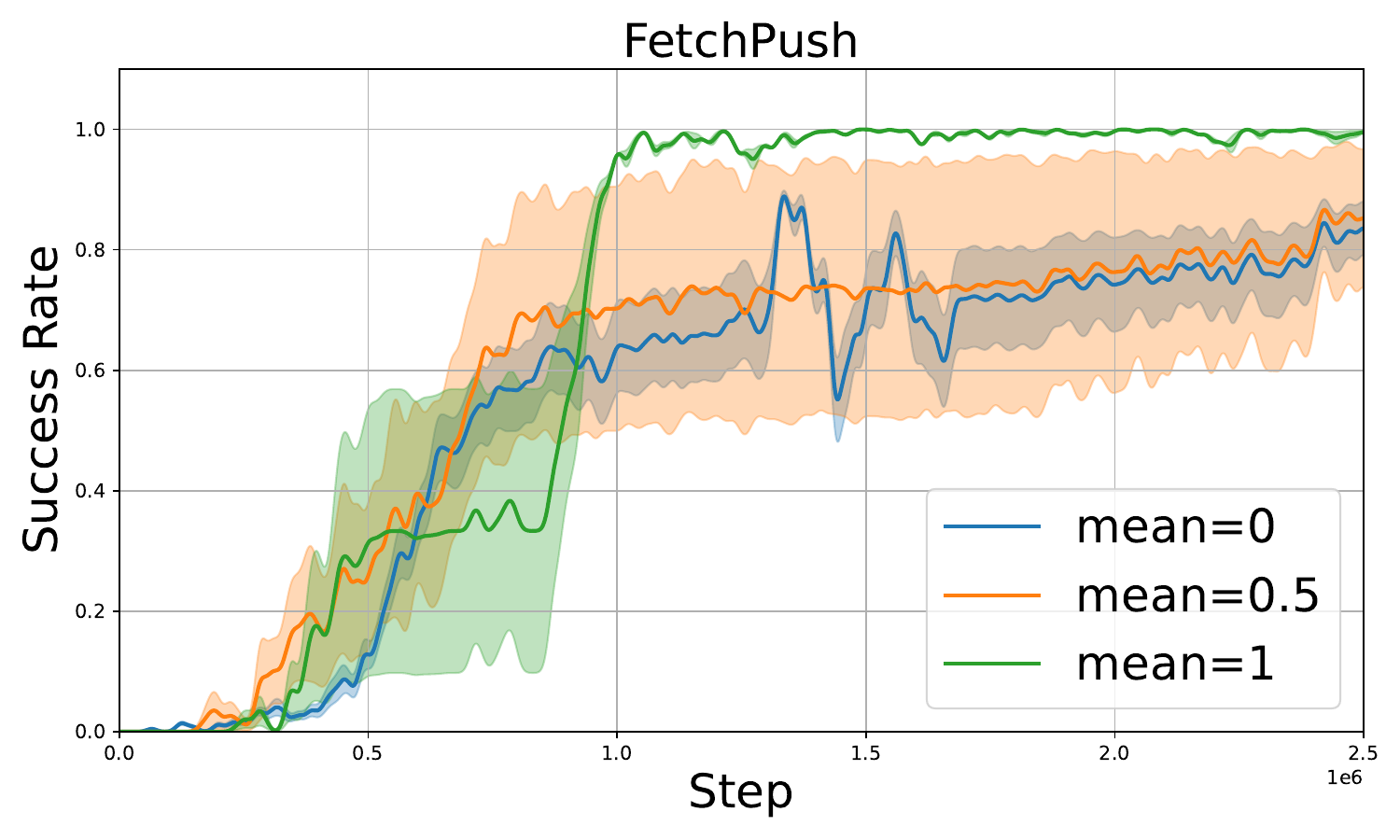} 
    \captionsetup{font=scriptsize}
    
  \end{subfigure}
  \hspace{0.1cm}
  \begin{subfigure}{0.15\textwidth}
    \centering
    \includegraphics[width=1.15\linewidth]{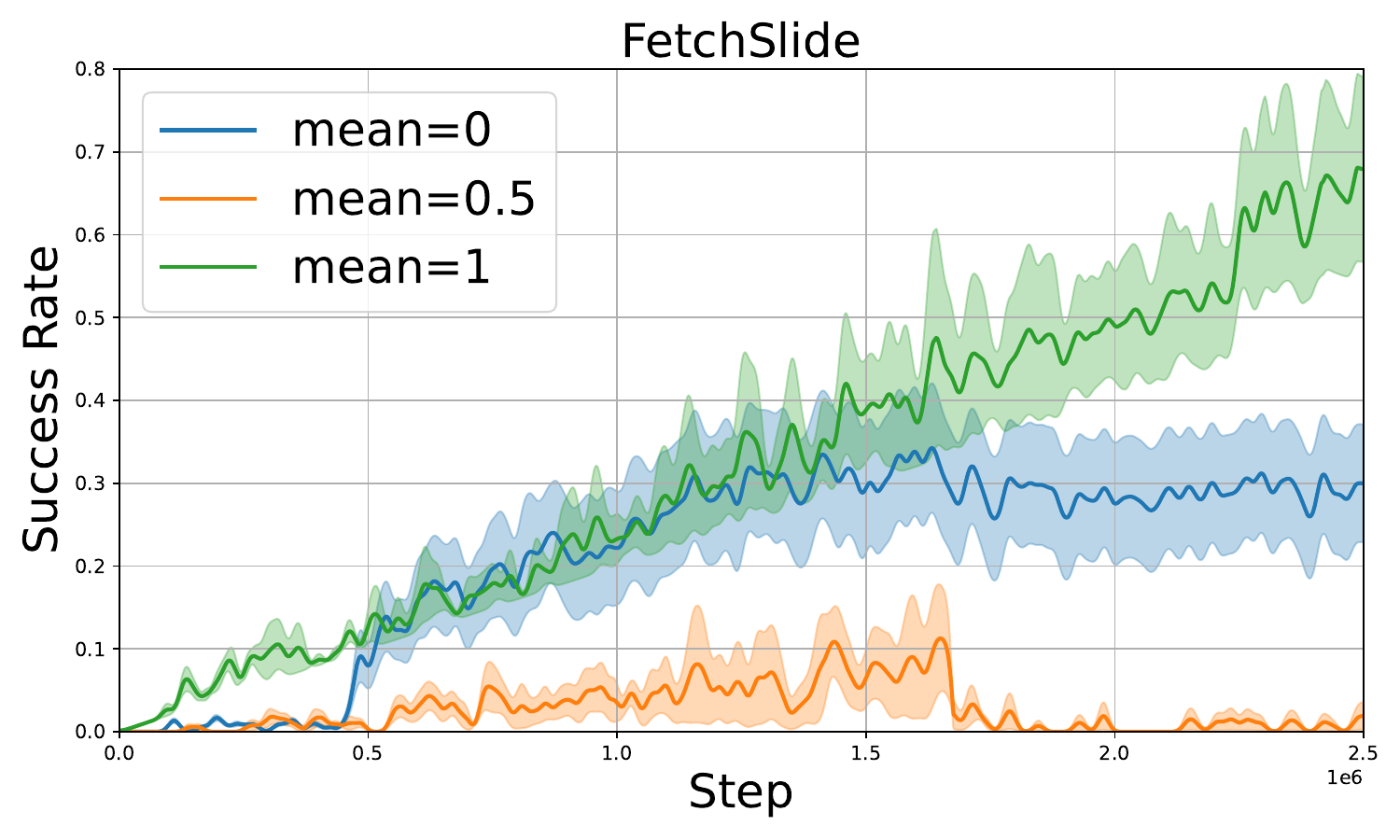} 
    \captionsetup{font=scriptsize}
  \end{subfigure}\\
  
      \begin{subfigure}{0.15\textwidth}
    \centering
    \includegraphics[width=1.15\linewidth]{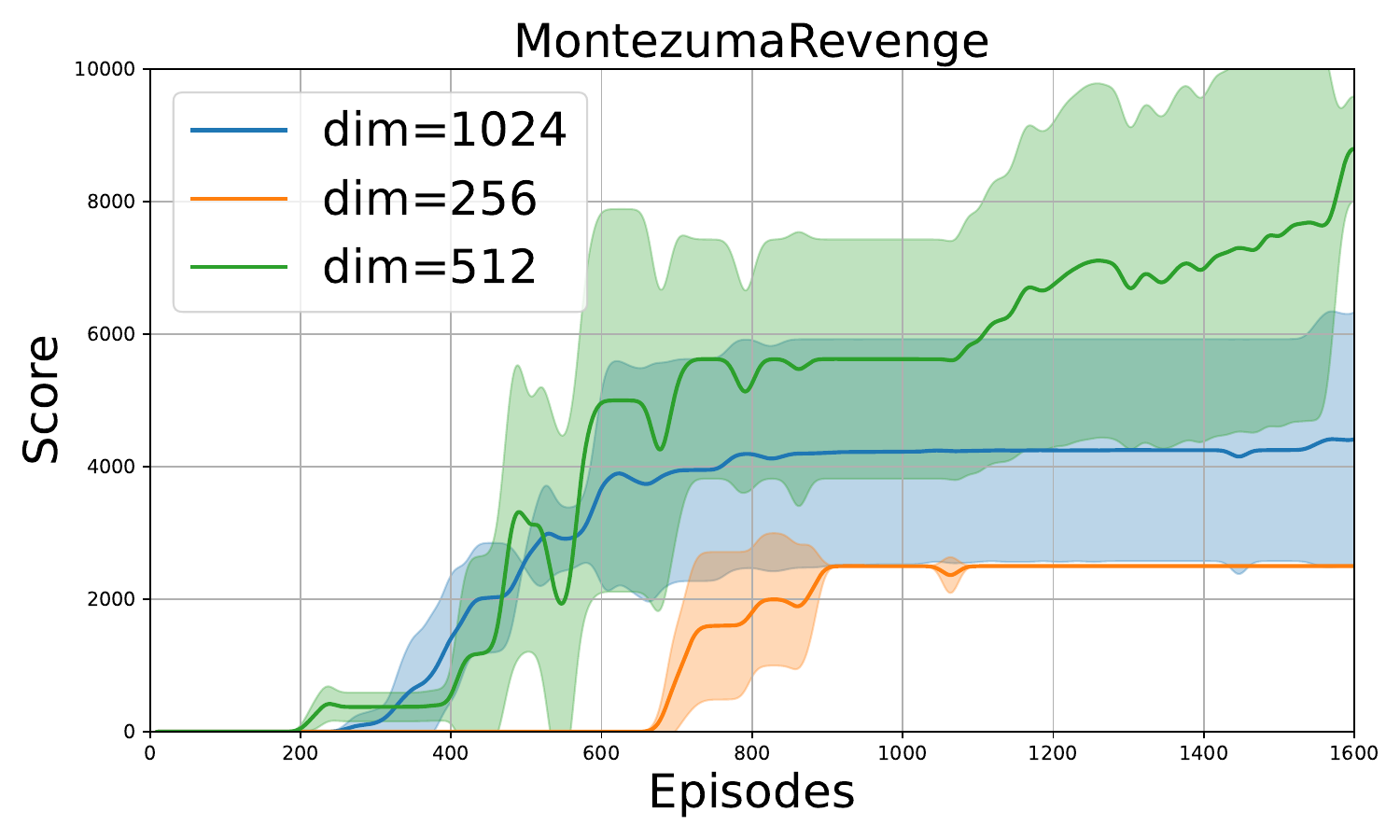}
    \captionsetup{font=scriptsize}
  \end{subfigure}
  \hspace{0.1cm}
  \begin{subfigure}{0.15\textwidth}
    \centering
    \includegraphics[width=1.15\linewidth]{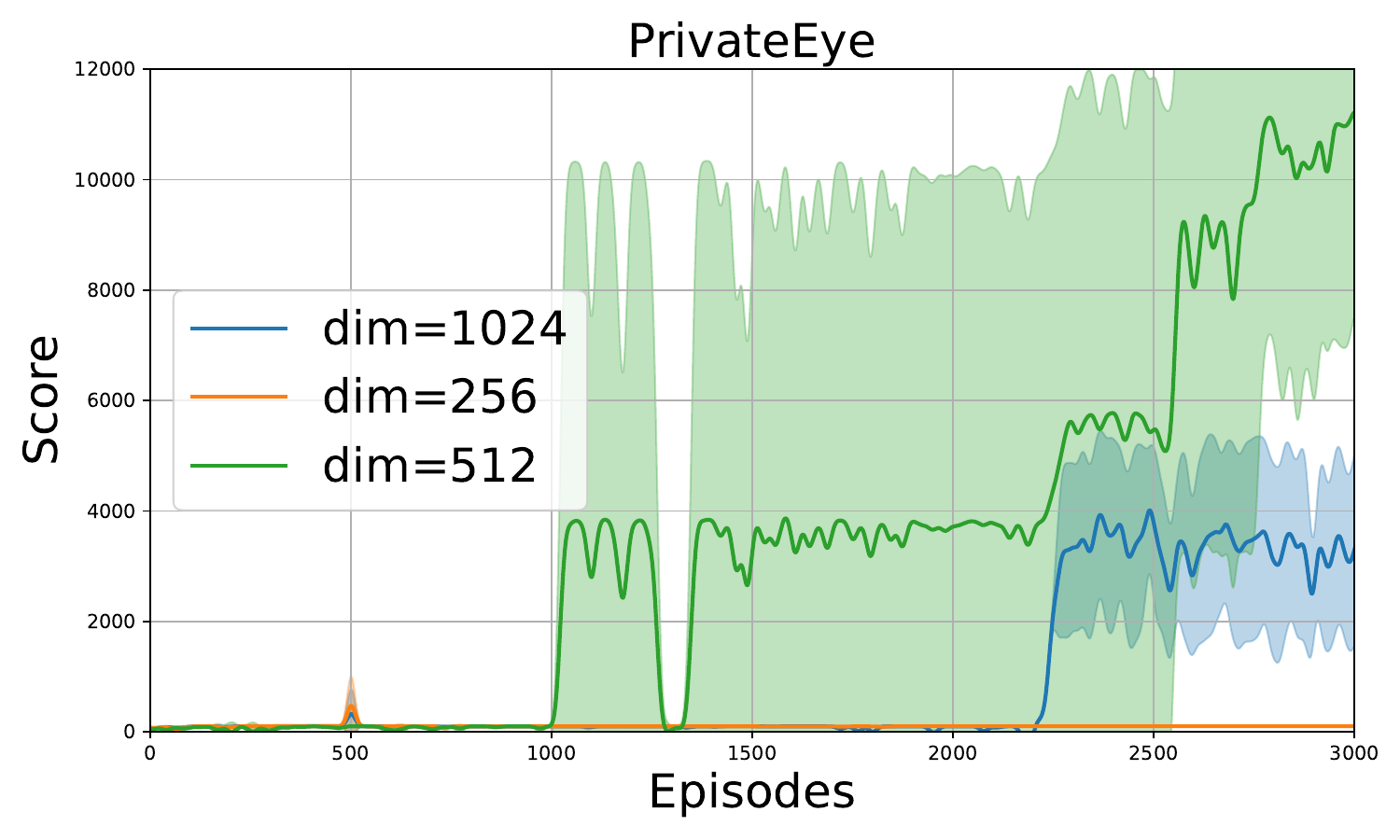}
    \captionsetup{font=scriptsize}
  \end{subfigure}
  \hspace{0.1cm}
  \begin{subfigure}{0.15\textwidth}
    \centering
    \includegraphics[width=1.15\linewidth]{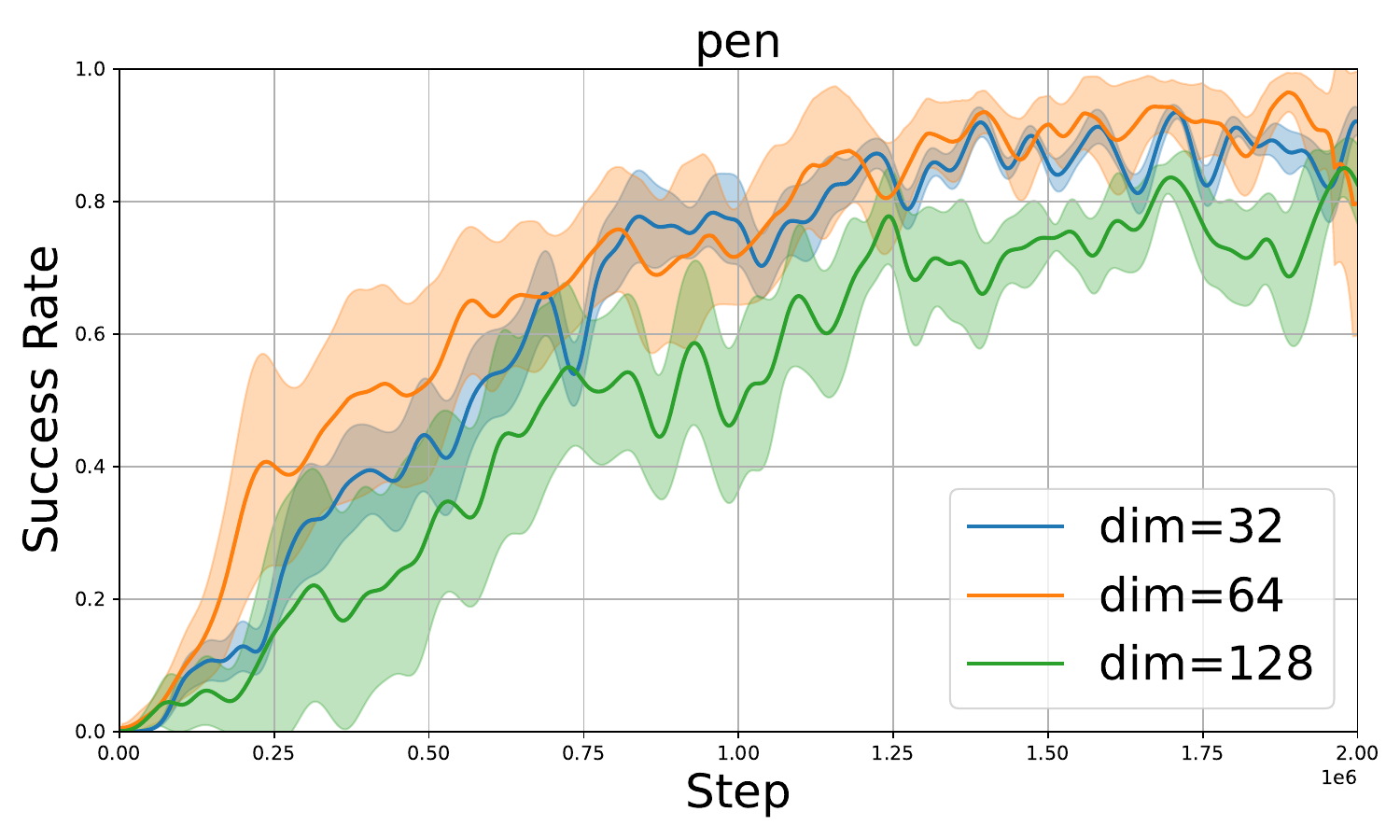} 
    \captionsetup{font=scriptsize}
  \end{subfigure}
  \hspace{0.1cm}
  \begin{subfigure}{0.15\textwidth}
    \centering
    \includegraphics[width=1.15\linewidth]{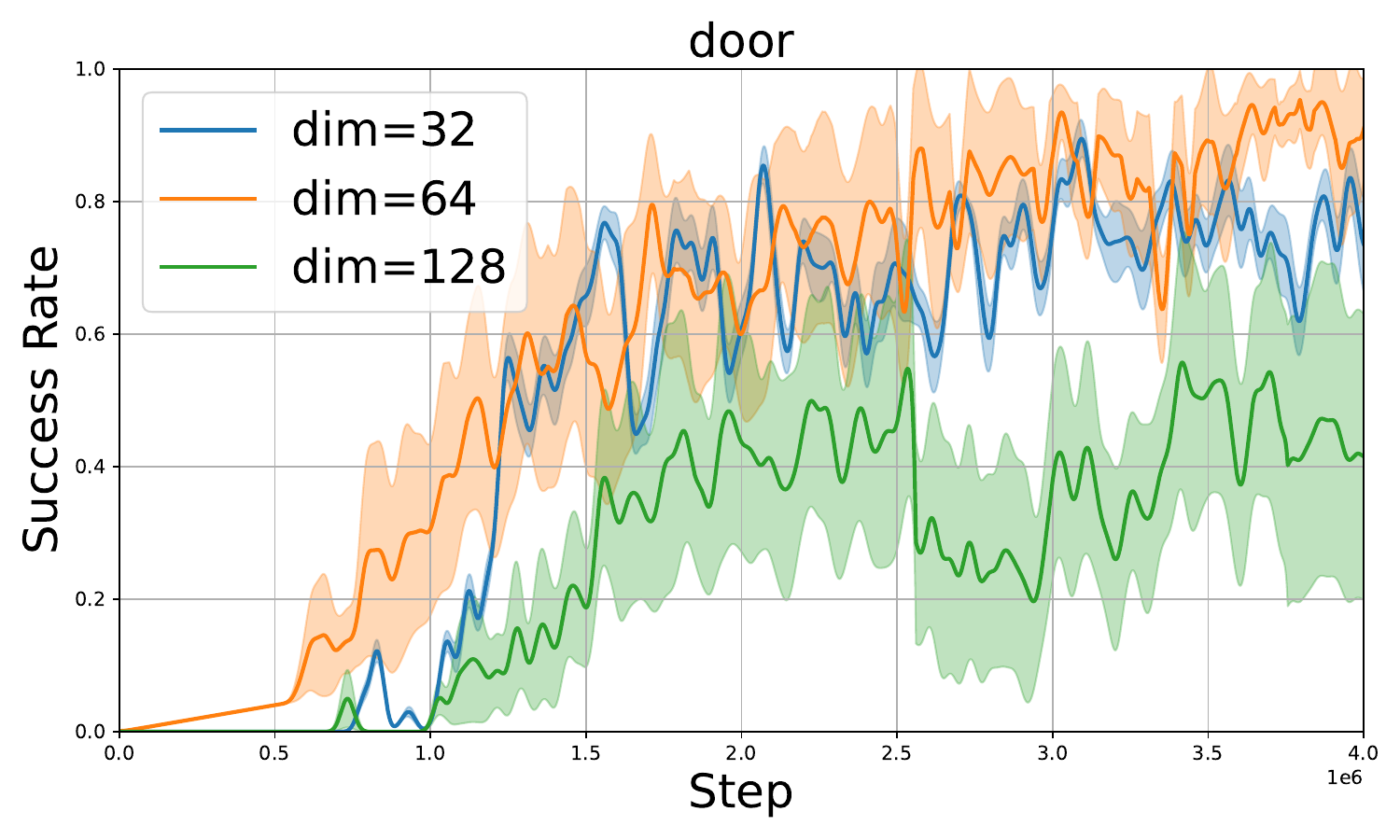} 
    \captionsetup{font=scriptsize}
  \end{subfigure}
  \hspace{0.1cm}
    \begin{subfigure}{0.15\textwidth}
    \centering
    \includegraphics[width=1.15\linewidth]{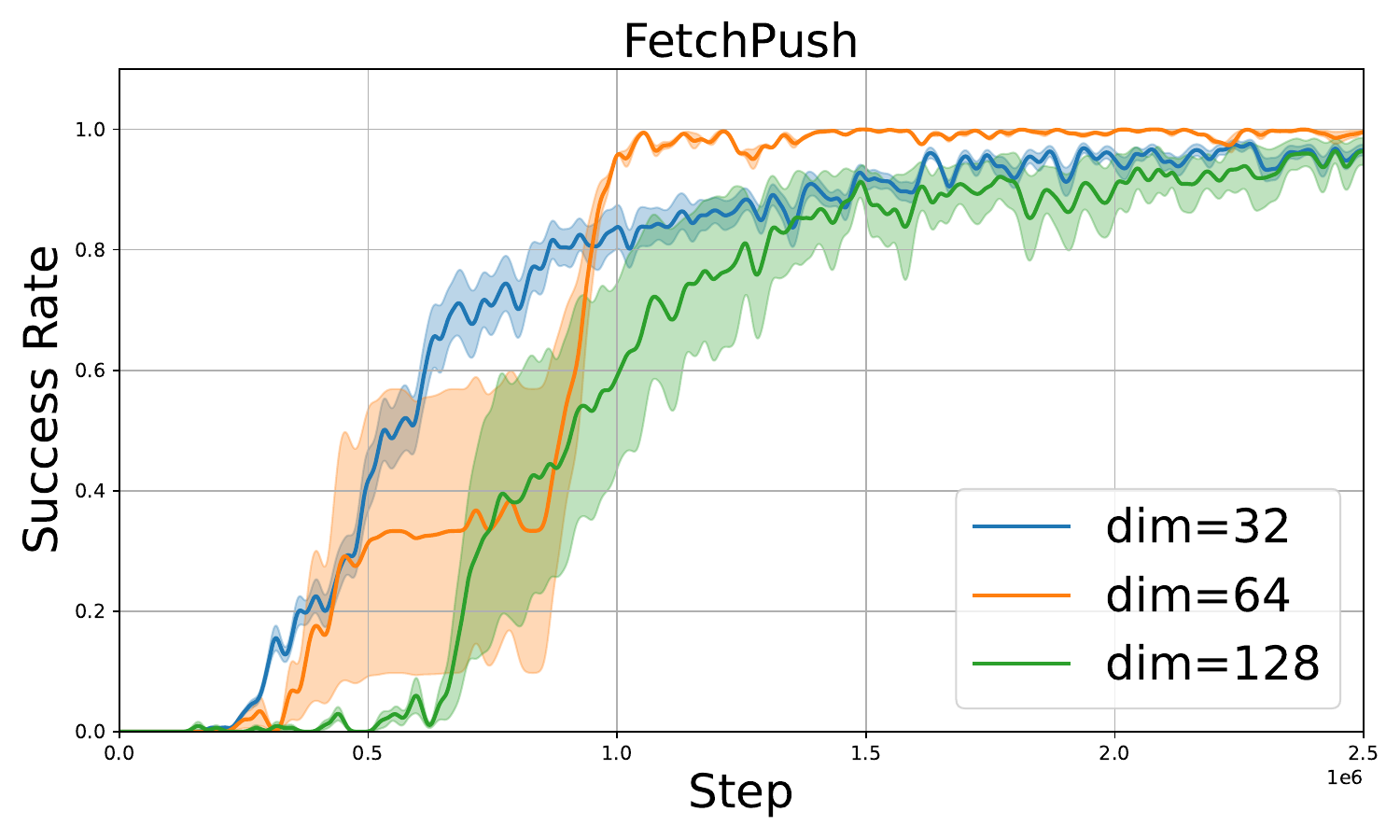} 
    \captionsetup{font=scriptsize}
  \end{subfigure}
  \hspace{0.1cm}
  \begin{subfigure}{0.15\textwidth}
    \centering
    \includegraphics[width=1.15\linewidth]{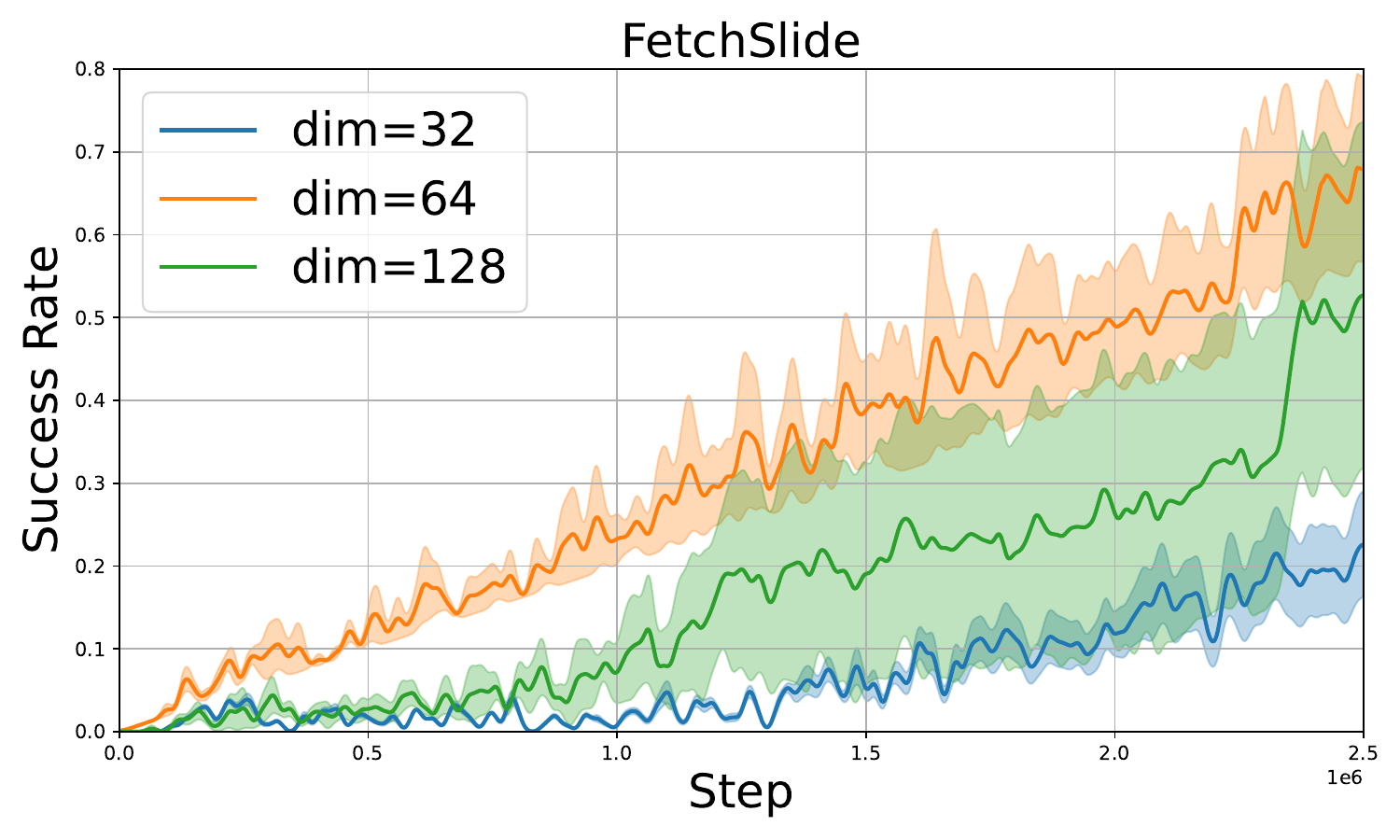} 
    \captionsetup{font=scriptsize}
  \end{subfigure}
\caption{We mainly analyzed the mean and variance of the target network, as well as the influence of the output dimensions in different environments on the exploration effect.}
\end{figure}

\section{Conclusions}
In this paper, we propose a simple and effective method for the exploration problem in online reinforcement learning: Random Distribution Distillation. We adopt the technical route based on the prediction-error method, by constructing prediction networks and target networks, and sampling the output of a target network from a normal distribution to promote exploration. Different from the RND method, we analyze the final bonus by constructing statistics to alleviate the problem of inconsistent bonus, and at the same time, we can conduct rigorous quantitative analysis on how the rewards will converge.

\section{Limitations}
The paper assumes that the target network distributions for different states are independent normal distributions. In practice, however, the parameters of the target network distribution are estimated by neural networks, meaning the distributions across different states may not be truly independent—similar states may exhibit some degree of distributional similarity.

{
\bibliographystyle{IEEEtran}
\bibliography{reference}
}






\clearpage
\appendix
{\Large
\textsc{Appendix}}

\section{Proof of Lemma \ref{Lemma1}}
\label{proof_A}
First, we prove that the expected value of \( z_n(s) \) is \( 1/n \). Since each $f_{\text{tar}}^i(s)$ is sampled independently from the distribution $\mathcal{N}(\mu_{\bar\varphi}(s),\sigma^2_{\bar\phi}(s))$, the expectation of $z_n(s)$ conditioned on $f_{\text{tar}}(s)$ is:
\begin{align}\label{mu1}
&\mathbb{E}[z_n(s)] = \frac{1}{\sigma^2_{\bar\phi}(s)}\mathbb{E}
\left[[f_{\theta^*}(s)]^2-2f_{\theta^*}(s)\mu_{\bar\theta}(s)+[\mu_{\bar\theta}(s)]^2\right] 
\nonumber\\ &= \frac{1}{\sigma^2_{\bar\phi}(s)}  \left[ \frac{\mathbb{E}[\sum_{i=1}^{n}f^i_{\text{tar}}(s)]}{n^2} - 2\mathbb{E}\left[\frac{1}{n} \sum_{i=1}^{n}f_{\text{tar}}^i(s)\right]\mu_{\bar\theta}(s) + [\mu_{\bar\theta}(s)]^2\right] \nonumber\\
& =\frac{1}{\sigma^2_{\bar\phi}(s)} \left[\frac{n\sigma^2_{\bar\phi}(s)+n[\mu_{\bar\theta}(s)]^2+n(n-1)[\mu_{\bar\theta}(s)]^2}{n^2} - [\mu_{\bar\theta}(s)]^2\right] = \frac{1}{n}.
\end{align}

Now let's analyze the variance of the $z_n(s)$. Since $f_\text{tar}$ follows a normal distribution, for the convenience of the derivation of the subsequent formula, we first provide some formulas for calculating the origin moments of a normal distribution as follows:
\begin{equation}\label{b2}
B_{2}(s) = [\mu_{\bar\theta}(s)]^2 + \sigma^2_{\bar\phi}(s).
\end{equation}
\begin{equation}\label{b3}
B_{3}(s) = [\mu_{\bar\theta}(s)]^3 + 3[\mu_{\bar\theta}.(s)]\sigma^2_{\bar\phi}(s).
\end{equation}
\begin{equation}\label{b4}
B_{4}(s) = [\mu_{\bar\theta}(s)]^4 + 6[\mu_{\bar\theta}(s)]^2\sigma^2_{\bar\phi}(s) + 3\sigma^4_{\bar\phi}(s).
\end{equation}

Similar to DRND, we have:
\begin{align}
&\mathbb{E}\left[[f_{\theta^*}(s)]^4\right]=  \frac{1}{n^{4}} \mathbb{E}\left[\sum_{i=1}^{n} f_\text{tar}^i(s)\right]^{4} \nonumber\\
= & \frac{1}{n^{4}}\left(\mathbb{E}\left[\sum_{i=1} [f_\text{tar}^i(s)]^{4}\right]+4 \mathbb{E}\left[\sum_{i \neq j} [f_\text{tar}^i(s)]^3 f_\text{tar}^j(s)\right]+3 \mathbb{E}\left[\sum_{i \neq j} [f_\text{tar}^i(s)]^2[f_\text{tar}^j(s)]^2\right]\right. \nonumber\\
& \left.+6 E\left[\sum_{i \neq j \neq k} f_\text{tar}^i(s) f_\text{tar}^j(s) [f_\text{tar}^k(s)]^{2}\right]+\mathbb{E}\left[\sum_{i \neq j \neq k \neq l} f_\text{tar}^i(s) f_\text{tar}^j(s) f_\text{tar}^k(s) f_\text{tar}^l(s)\right]\right) \nonumber\\
= & \frac{n B_{4}(s)+4 A_{n}^{2} \mu_{\bar\theta}(s) B_{3}(s)+3 A_{n}^{2} B_{2}^{2}(s)+6 A_{n}^{3} [\mu_{\bar\theta}(s)]^2 B_{2}(s)+A_{n}^{4} \mu^{4}(s)}{n^{4}} \nonumber\\
& \left(A_{n}^{i}=\frac{n!}{(n-i)!}\right)
\nonumber\\  = & \frac{n^2[\mu_{\bar\theta}(s)]^4+6n[\mu_{\bar\theta}(s)]^2\sigma^2_{\bar\phi}(s)+3\sigma^4_{\bar\phi}(s)}{n^2}.
\end{align}

\begin{align}
&\mathbb{E}\left[[f_{\theta^*}(s)]^3\right]=  \frac{1}{n^{4}} \mathbb{E}\left[\sum_{i=1}^{n} f_\text{tar}^i(s)\right]^{3} \nonumber\\
= & \frac{1}{n^{3}}\left(\mathbb{E}\left[\sum_{i=1} f_\text{tar}^i(s)^{3}\right]+3 \mathbb{E}\left[\sum_{i \neq j} [f_\text{tar}^i(s)]^2 f_\text{tar}^j(s)\right]+\mathbb{E}\left[\sum_{i \neq j \neq k} f_\text{tar}^i(s) f_\text{tar}^j(s) f_\text{tar}^k(s)\right]\right) \nonumber\\
= & \frac{n[\mu_{\bar\theta}(s)]^3+3\sigma^2_{\bar\phi}(s)[\mu_{\bar\theta}(s)]}{n}.
\end{align}

\begin{align}
\mathbb{E}\left[[f_{\theta^*}(s)]^2\right] &= \frac{\mathbb{E}[[\sum_{i=1}^{n}f^i_{\text{tar}}(s)]^2]}{n^2} \nonumber\\
&= \frac{n\sigma^2_{\bar\phi}(s)+n[\mu_{\bar\theta}(s)]^2+n(n-1)[\mu_{\bar\theta}(s)]^2}{n^2} \nonumber\\
&= \frac{\sigma^2_{\bar\phi}(s)}{n} + [\mu_{\bar\theta}(s)]^2.
\end{align}

So: 
\begin{align}
&\text{Var}[z_n(s)] = \frac{1}{\sigma^4_{\bar\phi}(s)}\text{Var}
\left[[f_{\theta^*}(s)]^2-2f_{\theta^*}(s)\mu_{\bar\theta}(s)\right] 
\nonumber\\ &= \frac{1}{\sigma^4_{\bar\phi}(s)}\{\text{Var}\left[[f_{\theta^*}(s)]^2\right] + 4[\mu_{\bar\theta}(s)]^2\text{Var}\left[f_{\theta^*}(s)\right] -4\mu_{\bar\theta}(s)\text{Cov}[[f_{\theta^*}(s)]^2,f_{\theta^*}(s)]\} \nonumber\\
&= \frac{1}{\sigma^4_{\bar\phi}(s)}\{\mathbb{E}[[f_{\theta^*}(s)]^4] -\mathbb{E}^2[[f_{\theta^*}(s)]^2]  + 4[\mu_{\bar\theta}(s)]^2\mathbb{E}[[f_{\theta^*}(s)]^2]  \nonumber\\
&\quad \quad \quad \quad- 4[\mu_{\bar\theta}(s)]^2\mathbb{E}^2[f_{\theta^*}(s)]-4\mu_{\bar\theta}(s)\mathbb{E}[f_{\theta^*}(s)]^3]+ 4[\mu_{\bar\theta}(s)]^2\mathbb{E}[[f_{\theta^*}(s)]^2]\} \nonumber\\
&= \frac{2}{n^2}.
\end{align}


\section{Proof of Corollary \ref{coro}}
\label{proof_B}
As noted in Appendix A of DRND \cite{yang2024exploration}, when the target network $f_{\text{tar}}^i(s) \sim \mathcal{N}(\mu_{\bar\theta}(s),\sigma^2_{\bar\phi}(s)$), the mean and variance of DRND statistic $y_n(s)$ are given by:
\begin{align}
\mathbb{E}[y_n(s)]  = \mathbb{E}[z_n(s)] = \frac{1}{n}.
\end{align}
\begin{align}
    &\text{Var}[y_n(s)]  = \frac{\text{Var}[[f_{\theta^*}(s)]^2]}{(B_2(s) - \mu_{\bar\theta}(s)^2)^2}\nonumber\\ & = \frac{\mathbb{E}[[f_{\theta^*}(s)]^4]-\mathbb{E}^2[[f_{\theta^*}(s)]^2]}{\sigma^2_{\bar\phi}(s)^4} \nonumber\\
    & = \frac{n^2[\mu_{\bar\theta}(s)]^4+6n[\mu_{\bar\theta}(s)]^2\sigma^2_{\bar\phi}(s)^2+3\sigma^2_{\bar\phi}(s)^4-\sigma^2_{\bar\phi}(s)^4-2n\sigma^2_{\bar\phi}(s)[\mu_{\bar\theta}(s)]^2-n^2[\mu_{\bar\theta}(s)]^4}{n^2\sigma^2_{\bar\phi}(s)^4}
    \nonumber\\ & = \frac{2}{n^2} + \frac{4[\mu_{\bar\theta}(s)]^2}{n\sigma^2_{\bar\phi}(s)}
    \nonumber\\ & \ge \text{Var}[z_n(s)].
\end{align}

\section{Proof of Theorem \ref{consistency}}
\label{proof_C}
For simplicity, we abbreviate $f^i_\text{tar}(s)$ as $f_i$, $\mu_{\bar \theta}(s)$ as $\mu$, $\sigma^2_{\bar \phi}(s)$ as $\sigma^2$ and $z_n(s)$ as $z_n$. Then: 
    \begin{align}
        z_n &=\frac{(\frac{1}{n}\sum_{i=1}^nf_i-\mu)^2}{\sigma^2}
        \nonumber\\&=\frac{1}{\sigma^2n^2}(\sum_{i=1}^nf_i-n\mu)^2
        \nonumber\\&=\frac{1}{\sigma^2n^2}[\sum_{i=1}^n(f_i-\mu)^2+2\sum_{i=1}^{n-1}\sum_{j>i}^n(f_i-\mu)(f_j-\mu)].
    \end{align}

    We first consider $X_n=\frac{1}{\sigma^2}\sum_{i=1}^n(f_i-\mu)^2=\sum_{i=1}^n(\frac{f_i-\mu}{\sigma})^2$. Since $f_i \sim \mathcal{N}(\mu,\sigma^2)$ and $f_i,f_j$ are independent when $i\ne j$, so $\frac{f_i-\mu}{\sigma}\sim \mathcal{N}(0,1)$ and $X_n=\sum_{i=1}^n(\frac{f_i-\mu}{\sigma})^2\sim \chi^2_n$. From the property of Chi-squared distribution, we know that $\mathbb{E}[X_n]=n$ and  $Var[X_n]=2n$.

    we can use Chebyshev's inequality as follows:
    \begin{align}
        p(|X_n-\mathbb{E}X_n|\ge \epsilon)&\le \frac{Var[X_n]}{\epsilon^2}
        \nonumber\\&=\frac{2n}{\epsilon^2}.
    \end{align}
    Using $\epsilon=n^2t$, we have:
    \begin{align}
 p(|X_n-\mathbb{E}X_n|\ge \epsilon)=p(|\frac{X_n}{n^2}-\frac{1}{n}|\ge t)\le \frac{2}{n^3t^2},
    \end{align}
    Taking $t=\sqrt{\frac{2}{\delta n^3}}$, we have:
    \begin{equation}
        p\left(|\frac{X_n}{n^2}-\frac{1}{n}|\ge \sqrt{\frac{2}{\delta n^3}}\right)\le \delta,
    \end{equation}
    for which we can conclude that $\frac{1}{\sigma^2 n^2} \sum_{i=1}^n (f_i - \mu)^2 \to \frac{1}{n}$.
    
    We then consider $Y_n=\sum_{i=1}^{n-1}\sum_{j>i}^n\frac{f_i-\mu}{\sigma}\frac{f_j-\mu}{\sigma}$. Since $\frac{f_i-\mu}{\sigma}$ and $\frac{f_j-\mu}{\sigma}$ are independent and follow a standard normal distribution, then their product can be expressed as $Z=UV \quad U,V\sim\mathcal{N}(0,1)$. The pdf of $Z$ is:
    \begin{align}
        f_Z(z) &= \int_{-\infty}^\infty f_U(x)f_V(\frac{z}{x})\frac{1}{|x|}dx 
        \nonumber\\&=\frac{1}{2\pi}\int_{-\infty}^\infty e^{-\frac{x^2}{2}}e^{-\frac{z^2}{2x^2}}\frac{1}{|x|}dx
        \nonumber\\&=\frac{1}{\pi}\int_{0}^\infty e^{-\frac{1}{2}(x^2+\frac{z^2}{x^2})}\frac{1}{x}ds
        \nonumber\\&=\frac{1}{2\pi}\int_{0}^\infty e^{-\frac{1}{2}(t+\frac{z^2}{t})}\frac{1}{t}dt.\quad\quad (t=x^2)
    \end{align}
    Note that the modified Bessel function of the second kind has the following expression: 
    \begin{align}
        2\left(\frac{b}{a}\right)^{\nu / 2} K_{\nu}(2 \sqrt{a b}) = \int_{0}^{\infty} e^{-(a t+b / t)} t^{\nu-1} d t.
    \end{align}
    When $a=\frac{1}{2},b=\frac{z^2}{2},v=0$, the expression can be simplified as:
    \begin{align}
    \int_{0}^{\infty} e^{-\left(t / 2+z^{2} /(2 t)\right)} \frac{1}{t} d t=2 K_{0}(|z|).
    \end{align}
    So $f_Z(z)=\frac{1}{2\pi}2K_0(|z|)=\frac{K_0(|z|)}{\pi}$, where $K_0$ denotes the second kind of modified Bessel function.

    The character function of $Z$ can be expressed as:
    \begin{align}
\phi(t)&=\int_{-\infty}^\infty e^{itz}\frac{K_0(z)}{\pi}dz
\nonumber\\&=\frac{2}{\pi}\int_{0}^\infty e^{itz}K_0(z)dz
\nonumber\\&=\frac{2}{\pi}\int_{0}^\infty \cos(tz) K_0(z)dz
\nonumber\\&=\frac{2}{\pi}(\frac{\pi}{2}\frac{1}{\sqrt{1+t^2}})
\nonumber\\&=\frac{1}{\sqrt{1+t^2}}.
    \end{align}
    So the MGF of $Z$ is $M_z(t)=\phi(it)=\frac{1}{\sqrt{1-t^2}} \quad |t|<1$. 
    
    Let's back to $Y_n$. The expectation of $Y_n$ is $\mathbb{E}[Y_n]=\frac{n(n-1)}{2}\mathbb{E}Z=0$ and the MGF of $Y_n$ is $M_Y(t)=(1-t^2)^{-\frac{n(n-1)}{4}} \quad |t|<1$. Using the Chernoff bound, we get:
    \begin{align}
         p(Y_n-\mathbb{E}Y_n\ge \epsilon) &\le \min_{\lambda\ge 0}\frac{\mathbb{E}[e^{\lambda(Y_n-\mathbb{E}Y_n)}]}{e^{\lambda\epsilon}}
         \nonumber\\&= \min_{\lambda \ge 0}M_Y(\lambda)e^{-\lambda\epsilon}
         \nonumber\\&=\min_{0 \le\lambda \le 1}(1-\lambda^2)^{-\frac{n(n-1)}{4}}e^{-\lambda\epsilon}.
    \end{align}
    Consider $h(x)=(1-x^2)^{-\frac{n(n-1)}{4}}e^{-x\epsilon} \quad |x|<1$ and $\ln h(x)=-\frac{n(n-1)}{4}\ln(1-x^2)-x\epsilon \quad |x|<1$. Let $\frac{\mathrm{d}}{\mathrm{d}x}\ln h(x)=\frac{n(n-1)}{2}\frac{x}{1-x^2}-\epsilon=0$ we get $\frac{2 \epsilon n(n-1)}{n(n-1)} x^{2}+x-\frac{2 \epsilon}{n(n-1)}=0$. This is a quadratic equation and the solutions of the equation are: $x=\frac{-n(n-1)\pm \sqrt{n^2(n-1)^2+16 \epsilon^{2}}}{4 \epsilon}$. Since $x\in[-1,1]$, the only solution $x_0$ to minimize $h(x)$ is $x_0=\frac{-n(n-1)+ \sqrt{n^2(n-1)^2+16 \epsilon^{2}}}{4 \epsilon}$. So we have:
    \begin{align}
 &p(Y_n-\mathbb{E}Y_n\ge \epsilon)\le \min_{0 \le\lambda \le 1}(1-\lambda^2)^{-\frac{n(n-1)}{4}}e^{-\lambda\epsilon}
 \nonumber\\ &= \exp\{-\frac{n(n-1)}{4}\ln(1-x_0^2)-x_0\epsilon\}.
    \end{align}
    Taking $\epsilon=tn^2$, we have:
    \begin{align}
         &p(Y_n-\mathbb{E}Y_n\ge \epsilon)= p(\frac{Y_n}{n^2}-0\ge t)\nonumber\\& \le \exp\{-\frac{n(n-1)}{4}\ln(1-x_0^2)-x_0tn^2\}
    \nonumber\\ &\le \exp\{\frac{-n(n-1) + \sqrt{n^2(n-1)^2 + 16 t^2n^4}}{4}\}.
    \end{align}
    Similarly, since the MGF is symmetric in origin, we can get $p(\frac{Y_n}{n^2}\ge t)\le \exp\{\frac{-n(n-1) + \sqrt{n^2(n-1)^2 + 16 t^2n^4}}{4}\} \}$, so $p(|\frac{Y_n}{n^2}|\ge t)=p(|\frac{1}{n^2\sigma^2}\sum_{i=1}^{n-1}\sum_{j>i}^n(f_i-\mu)(f_j-\mu)|\ge t)\le 2\exp\{\frac{-n(n-1) + \sqrt{n^2(n-1)^2 + 16 t^2n^4}}{4}\}$. Taking $t=\frac{ \sqrt{ \ln\left( \frac{2}{\delta} \right) \left[ \frac{n(n-1)}{2} + \ln\left( \frac{2}{\delta} \right) \right] } }{n^2 }$, we have:
    \begin{equation}
        p\left(|\frac{Y_n}{n^2}|\ge \frac{ \sqrt{ \ln\left( \frac{2}{\delta} \right) \left[ \frac{n(n-1)}{2} + \ln\left( \frac{2}{\delta} \right) \right] } }{n^2 }\right)\le \delta,
    \end{equation}
    which indicates that $\frac{2}{\sigma^2n^2}\sum_{i=1}^{n-1}\sum_{j>i}^n(f_i-\mu)(f_j-\mu) \to 0$.

    In conclusion, we have:
    \begin{align}
        &p(|z_n-\frac{1}{n}|>t)= p(|\frac{1}{\sigma^2n^2}[\sum_{i=1}^n(f_i-\mu)^2+2\sum_{i=1}^{n-1}\sum_{j>i}^n(f_i-\mu)(f_j-\mu)]-\frac{1}{n}|>t)
        \nonumber\\&\le p(|\frac{1}{\sigma^2n^2}[\sum_{i=1}^n(f_i-\mu)^2-\frac{1}{n}|>t)+p(|\frac{2}{\sigma^2n^2}\sum_{i=1}^{n-1}\sum_{j>i}^n(f_i-\mu)(f_j-\mu)]-\frac{1}{n}|>t).
    \end{align}
    Taking $t=\sqrt{\frac{2}{\delta n^3}}+\frac{ \sqrt{ \ln\left( \frac{2}{\delta} \right) \left[ \frac{n(n-1)}{2} + \ln\left( \frac{2}{\delta} \right) \right] } }{n^2 }$, we have:
    \begin{align}
&p(|z_n-\frac{1}{n}|>t)\le p\left(|\frac{1}{\sigma^2n^2}[\sum_{i=1}^n(f_i-\mu)^2-\frac{1}{n}|>t\right)\nonumber\\
&+p\left(|\frac{2}{\sigma^2n^2}\sum_{i=1}^{n-1}\sum_{j>i}^n(f_i-\mu)(f_j-\mu)]-\frac{1}{n}|>t\right)
\nonumber\\&< p\left(|\frac{1}{\sigma^2n^2}[\sum_{i=1}^n(f_i-\mu)^2-\frac{1}{n}|>\sqrt{\frac{2}{\delta n^3}}\right)
\nonumber\\&+ p\left(\frac{2}{\sigma^2n^2}\sum_{i=1}^{n-1}\sum_{j>i}^n(f_i-\mu)(f_j-\mu)]-\frac{1}{n}|>\frac{ \sqrt{ \ln\left( \frac{2}{\delta} \right) \left[ \frac{n(n-1)}{2} + \ln\left( \frac{2}{\delta} \right) \right] } }{n^2 }\right)
\nonumber\\&\le \delta + \delta
\nonumber\\&=2\delta.
    \end{align}
So 
\begin{align}
p\left(|z_n-\frac{1}{n}|>\sqrt{\frac{2}{\delta n^3}}+\frac{ \sqrt{ \ln\left( \frac{2}{\delta} \right) \left[ \frac{n(n-1)}{2} + \ln\left( \frac{2}{\delta} \right) \right] } }{n^2 }\right)<2\delta,
\end{align}
which means that $z_n \to \frac{1}{n}$ and our proof is complete.

\section{Implementation Details and Experimental Settings}
\label{details}
Our experiments were performed by using the following hardware and software:
\begin{itemize}[noitemsep,leftmargin=*]
\item \textit{GPUs} NVIDIA GeForce RTX 4090
\item \textit{Python} 3.10.0
\item \textit{Numpy} 1.23.5
\item \textit{Gymnasium} 0.28.1
\item \textit{Gymnasium-robotics} 1.2.2
\item \textit{Pytorch} 1.13.0
\item \textit{MuJoCo-py} 2.1.2.14
\item \textit{MuJoCo} 2.3.3
\end{itemize}

\begin{figure}[htp]
  \centering
  \hspace{-0.2cm}
  \begin{subfigure}{0.16\textwidth}
    \centering
    \includegraphics[width=1\linewidth]{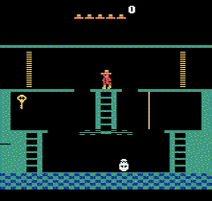}
    \captionsetup{font=scriptsize}
    \caption{Montezuma} 
    
  \end{subfigure}
  \begin{subfigure}{0.16\textwidth}
    \centering
    \includegraphics[width=1\linewidth]{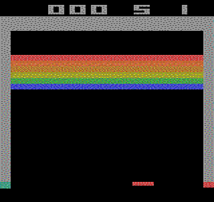}
    \captionsetup{font=scriptsize}
    \caption{Breakout} 
    
  \end{subfigure}
  \begin{subfigure}{0.16\textwidth}
    \centering
    \includegraphics[width=1\linewidth]{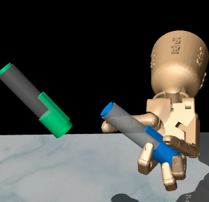} 
    \captionsetup{font=scriptsize}
    \caption{Adroit Pen} 
    
  \end{subfigure}
  \begin{subfigure}{0.16\textwidth}
    \centering
    \includegraphics[width=1\linewidth]{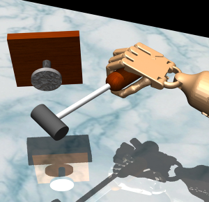} 
    \captionsetup{font=scriptsize}
    \caption{Adroit Hammer} 
    
  \end{subfigure}
    \begin{subfigure}{0.16\textwidth}
    \centering
    \includegraphics[width=1.06\linewidth]{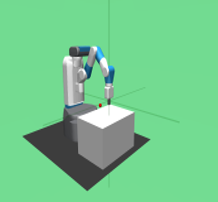} 
    \captionsetup{font=scriptsize}
    \caption{Fetch Reach} 
    
  \end{subfigure}
  \hspace{0.1cm}
  \begin{subfigure}{0.16\textwidth}
    \centering
    \includegraphics[width=1\linewidth]{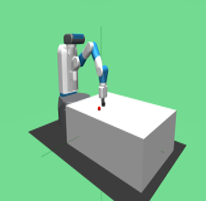} 
    \captionsetup{font=scriptsize}
    \caption{Fetch Slide} 
    
  \end{subfigure}\\

    \hspace{-0.2cm}
    \begin{subfigure}{0.16\textwidth}
    \centering
    \includegraphics[width=1\linewidth]{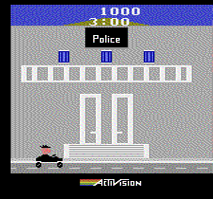}
    \captionsetup{font=scriptsize}
    \caption{PrivateEye} 
    
  \end{subfigure}
  \begin{subfigure}{0.16\textwidth}
    \centering
    \includegraphics[width=1\linewidth]{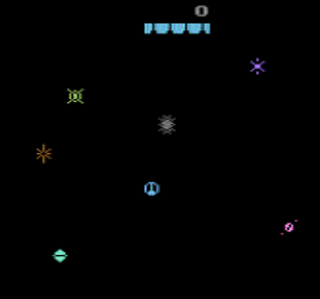}
    \captionsetup{font=scriptsize}
    \caption{Gravitar} 
    
  \end{subfigure}
  \begin{subfigure}{0.16\textwidth}
    \centering
    \includegraphics[width=1\linewidth]{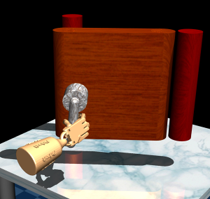} 
    \captionsetup{font=scriptsize}
    \caption{Adroit Door} 
    
  \end{subfigure}
  \begin{subfigure}{0.16\textwidth}
    \centering
    \includegraphics[width=0.96\linewidth]{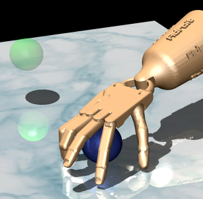} 
    \captionsetup{font=scriptsize}
    \caption{Adroit Relocate} 
    
  \end{subfigure}
  \hspace{-0.1cm}
    \begin{subfigure}{0.16\textwidth}
    \centering
    \includegraphics[width=1.08\linewidth]{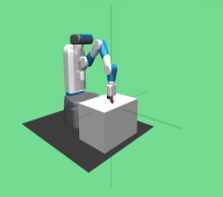} 
    \captionsetup{font=scriptsize}
    \caption{Fetch Push} 
    
  \end{subfigure}
  \hspace{0.1cm}
  \begin{subfigure}{0.16\textwidth}
    \centering
    \includegraphics[width=1\linewidth]{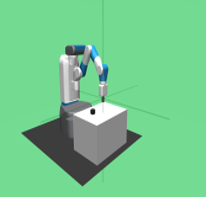} 
    \captionsetup{font=scriptsize}
    \caption{Fetch PickPlace} 
    
  \end{subfigure}
\caption{Environments of experiments}
\label{experiments}
\vspace{-0.8cm}
\end{figure}

In our online experimental evaluations, we utilized the ‘NoFrameskip-v4’ variant for the sparse reward Atari game environments to facilitate execution. These experiments were conducted across 128 parallel instances, adhering to the default configurations and network architecture as specified in the RND framework \cite{burda2018exploration}. For the Adroit and Fetch manipulation tasks, we employed the ‘v0’ version for the Adroit suite and the ‘v2’ version for the Fetch suite, respectively. Within the ‘Relocate’ task of the Fetch environment, we implemented an episode termination condition upon the ball exiting the table. Furthermore, task completion was defined as achieving a Euclidean distance of less than 0.2 between the target object and the designated goal point. These tasks present a substantial learning challenge for conventional methodologies, primarily attributable to the characteristics of the dataset, which comprises limited human demonstrations within a sparse-reward, complex, and high-dimensional robotic manipulation paradigm \cite{lyu2022double}. Consistent with the observations of \cite{lobel2022optimistic}, we opted not to incorporate random state restarts, as such interventions may diminish the inherent requirement for effective exploration strategies. For the specification of goal locations in non-default task versions, we adopted the settings outlined by \cite{lobel2023flipping}.

we provide the detailed configurations of the tasks in the experiments, which are shown in Figure \ref{experiments}.

\begin{itemize}[noitemsep,leftmargin=*]
    \item \textit{FetchReach}: The robot arm is trained to reach a target position. The target position is randomly generated in the workspace, and the reward is given if the robot reaches the target position. Maximum episode length is $2.5\times10^6$ steps.
    \item \textit{FetchPush}: The robot arm is trained to push an object to a target position. The target position is randomly generated on the table, and the reward is given if the object reaches the target position. Maximum episode length is $2.5\times10^6$ steps.
    \item \textit{RobotSlide}: The robot arm is trained to slide an object to a target position. The target position is generated randomly on the table, and the reward is given if the object reaches the target position. Maximum episode length is $2.5\times10^6$ steps.
    \item \textit{RobotPickPlace}: The robot arm is trained to pick and place an object to a target position. The target position is randomly generated in the space, and the reward is given if the object reaches the target position. Maximum episode length is $2.5\times10^6$ steps.
    \item \textit{MontezumaRevenge}: The agent is trained to navigate through a series of rooms to collect keys and reach the final room. The reward is given if the agent successfully reaches one new room. Maximum episode length is 1600 episodes.
    \item \textit{PrivateEye}: The agent navigates a map as a detective, collecting clues (reward: based on collecting clues and progressing the investigation; goal: solve the mystery, often implicitly defined by game progression). Maximum episode length is 3000 episodes.
    \item \textit{Breakout}: The agent controls a paddle to bounce a ball and break a wall of bricks (reward: for hitting and breaking bricks; goal: clear all the bricks or achieve a high score). Maximum episode length is 5000 episodes.
    \item \textit{Gravitar}: The agent pilots a spacecraft through hostile environments, destroying enemies and navigating gravity (reward: for destroying enemies, collecting fuel, and surviving; goal: progress through the levels and survive). Maximum episode length is 5000 episodes.
    \item \textit{Pen}: The agent manipulates a robotic hand to follow a target pen trajectory (reward: dense shaping based on trajectory tracking error; goal: minimize tracking error over the episode). Maximum episode length is $2.5\times10^6$ steps.
    \item \textit{Hammer}: The agent controls a robotic hand to hammer a nail to a target depth (reward: sparse, primarily at the end based on nail depth; goal: drive the nail head to the desired depth). Maximum episode length is $3.5\times10^6$ steps.
    \item \textit{Door}: The agent manipulates a robotic hand to open a hinged door (reward: sparse, primarily upon successfully opening the door; goal: achieve a certain open angle of the door). Maximum episode length is $4.5\times10^6$ steps.
    \item \textit{Relocate}: The agent controls a robotic hand to pick up and move an object to a target location on a table (reward: sparse, primarily upon the object reaching the target; goal: position the object within a threshold distance of the target). Maximum episode length is $4.5\times10^6$ steps.
\end{itemize}

Furthermore, we also provide the detailed dimensions of the states (observation space) and action space in our evaluated tasks in Table \ref{tab:appendix-state-dimensions} and Table~\ref{tab:appendix-action-dimensions}.

\begin{table*}[h]
\centering
\small
\caption{The dimensions of the states in the evaluated tasks.}
\label{tab:appendix-state-dimensions}
\begin{tabular}{cc}
    \toprule
    Domain (Tasks) & Dimension \\
    \midrule
    Adroit robot (\textit{Door}, \textit{Relocate}) & 39 \\
    Adroit robot (\textit{Pen}, \textit{Hammer}) & 45, 46 \\
    \textit{FetchReach} & 20 \\
    \textit{FetchPush}, \textit{FetchSlide} and \textit{FetchPickPlace} & 35 \\
    Atari games  & $84\times84 = 7056$ \\
    \textit{MountainCar} & 2 \\
    \bottomrule
\end{tabular}
\end{table*}

\begin{table*}[h]
\centering
\small
\caption{The dimensions of the action in the evaluated tasks.}
\label{tab:appendix-action-dimensions}

\begin{tabular}{cc}
    \toprule
    Domain (Tasks) & Dimension \\
    \midrule
    Adroit robot (\textit{Pen}, \textit{Hammer}, \textit{Door} and \textit{Relocate}) & 24, 26, 28, 30 \\
    \textit{FetchReach}, \textit{FetchPush}, \textit{FetchSlide} and \textit{FetchPickPlace} & 4 \\
    Atari games & 18 \\
    \textit{MountainCar} & 3 \\
    \bottomrule
\end{tabular}
\end{table*}

\section{Hyperparameters}
\label{hyper}
The hyperparameters are shown in Table \ref{hyperparameters} in online experiments. We employ distinct parameters and networks for Atari games and continuous control environments because Atari game observations are images, while observations for Adroit and Fetch tasks consist of states.

\begin{table*}[h]
  \centering
  \caption{Hyperparameter Settings}
  \label{hyperparameters}
  \begin{tabular}{l|l|l}
    \toprule
    \textbf{Name} & \textbf{Description} & \textbf{Value} \\
    \midrule
    $lr_{actor}$ & learning rate of the actor network & 3e-4 (1e-4 on Atari) \\
    $lr_{critic}$ & learning rate of the critic network & 3e-4 (1e-4 on Atari) \\
    $lr_{rdd}$ & learning rate of the RDD network & 3e-4 (1e-4 on Atari) \\
    optimizer & type of optimizer & Adam \\
    $\tau$ & soft update rate & 0.005 \\
    $\gamma$ & discount return & 0.99 \\
    $\lambda_{\text{GAE}}$ & coefficient of GAE & 0.95 \\
    $\epsilon$ & PPO clip coefficient & 0.1 \\
    $M$ & number of environments & 128 \\
    $h$ & number of hidden layer dimensions & 64 (512 on Atari) \\
    $e$ & number of RDD output dimensions & 64 (512 on Atari) \\
    $f$ & activation function & ReLU \\
    $K$ & number of training epochs & 4 \\
    $\beta$ & coefficient of intrinsic reward & 1 \\
    $\mu$ & The mean of the output by the target network & 1 \\
    $\sigma$ & The variance of the output by the target network & 1.0 \\
    \bottomrule
  \end{tabular}
\vspace{-0.7cm}
\end{table*}

\section{Diagram of RDD}
\label{Diagram}
\begin{figure}[h]
\centering
\vspace{-0.7cm}
\includegraphics[width=0.8\linewidth]{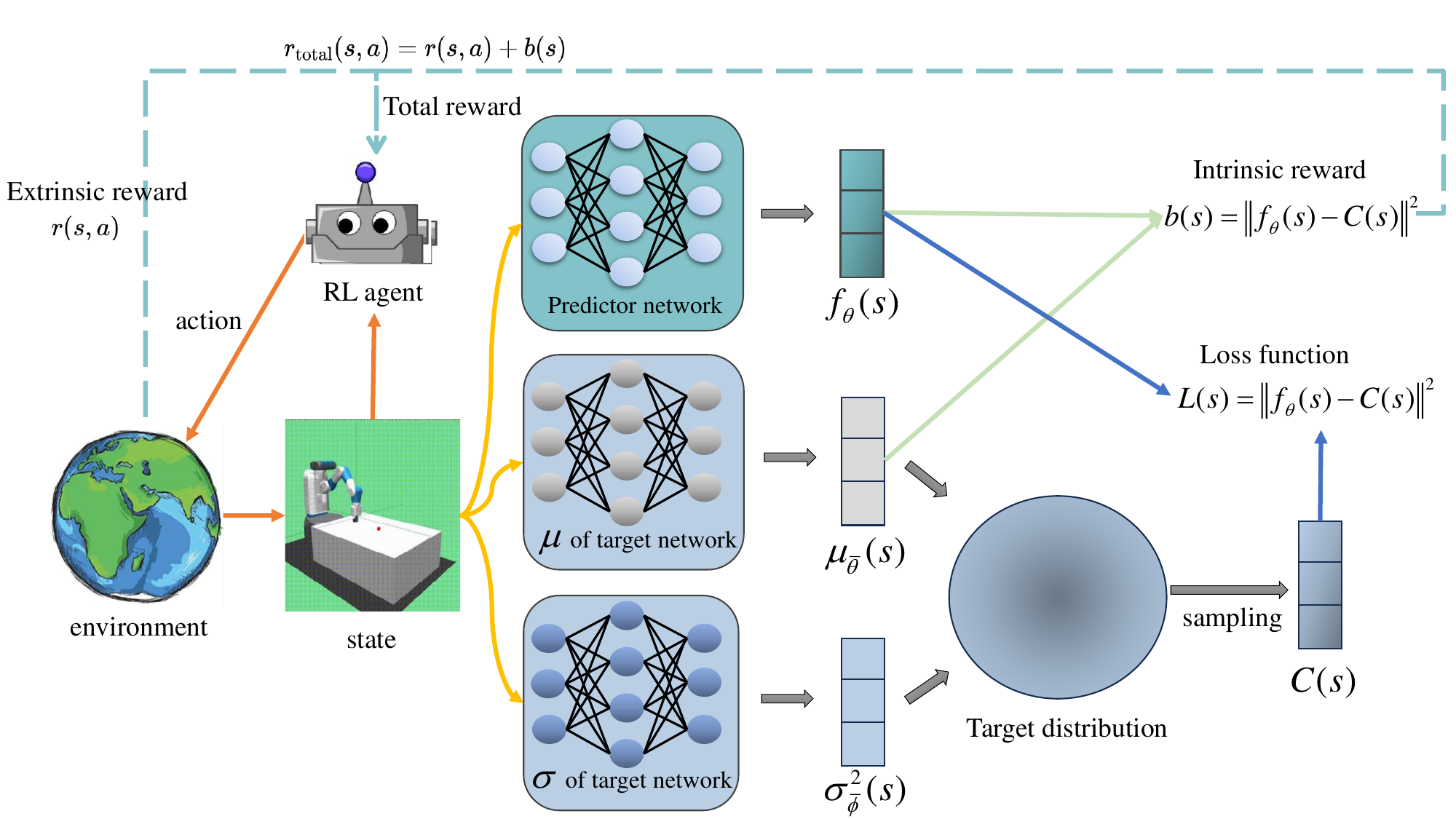} 
\caption{Diagram of RDD}
\label{rdd_plot}
\end{figure}

\section{Compute Resources}
\label{sec:appendix-resources}

The experiments in this paper were conducted on a computing cluster, with the detailed hardware configurations listed in Table~\ref{tab:appendix-resources}. The computing time for the RDD algorithm in each task (steps or episodes for experiments in Section) was approximately $3 \pm 2$ hours for Fetch robot tasks, $8 \pm 2$ hours for Adroit robot tasks and 1.5 days for Atari tasks.

\begin{table*}[h]
\centering
\vspace{-0.7cm}
\small
\caption{The compute resources used in the experiments}
\label{tab:appendix-resources}
\begin{tabular}{cc}
    \toprule
    Component & Specification \\
    \midrule
    Operating System (OS) & Ubuntu 20.04 \\
    Central Processing Unit (CPU) & 2x Intel Xeon Gold 6326 \\
    Random Access Memory (RAM) & 512GB \\
    Graphics Processing Unit (GPU) & 8x NVIDIA GeForce RTX 4090 24GB \\
    Brand & Supermicro 2022 \\
    \bottomrule
\end{tabular}
\vspace{-0.7cm}
\end{table*}

\section{RDD Pseudo-code}
\label{Pseudo-code}
We have listed the pseudo-code of our method as shown in Algorithm \ref{Pseudo-code1}.
    \begin{algorithm}
        \caption{PPO-RDD online pseudo-code}
        \label{Pseudo-code1}
        \begin{algorithmic}[1]
            \REQUIRE Number of training steps $M$, number of update steps $K$, parameters $\bar{\theta}$ and $\bar{\varphi}$ of $\mu$ and $\sigma$, scale of intrinsic reward $\lambda$
            \STATE Initialize policy parameters $\phi$
            \STATE Initialize Q-function parameters $\varphi$ and target Q-function parameters $\varphi'$
            \STATE Initialize predictor network parameters $f_\theta$ and target networks parameters $\mu_{\bar{\theta}}$ and $\sigma_{\bar{\varphi}}$ for $f_{tar}$
                \STATE Initialize replay buffer $D$
                \STATE $d \gets 0$, $t \gets 0$
                \STATE $s_0 = $ env.reset()
                \WHILE{not $d$}
                    \STATE $a_t \sim \pi(a_t|s_t)$
                    \STATE Rollout $a_t$ and get $(s_{t+1},r_t,d)$
                    \STATE Compute the mean $\mu_\theta(s_t)$ and second moment $\sigma_{\bar{\varphi}}(s_t)$
                    \STATE Compute intrinsic reward $b(s_{t+1},a_t)$ using $\mathrm{Eq.}$ \ref{bonus}
                    \STATE Add transition $(s_t,a_t,r_t,b(s_{t+1},a_t),s_{t+1})$ to $D$
                    \STATE $t \gets t+1$
                \ENDWHILE
                \STATE Normalize the intrinsic rewards contained in $D$
                \STATE Calculate returns $R_I$ and advantages $A_I$ for intrinsic reward
                \STATE Calculate returns $R_E$ and advantages $R_E$ for extrinsic reward
                \STATE Calculate combined advantages $A$ = $R_I$ + $R_E$
                \STATE $\phi_{old} \gets \phi$
                \FOR{$j=1:K$}
                   
                    \STATE Update $\phi$ with gradient ascent using 
                    \[
\scalebox{0.9}{
    $\nabla_\phi \frac{1}{|D|} \sum_{D}^{} \min \left(\frac{\pi_{\phi}(a \mid s)}{\pi_{\phi_{\text{old}}}(a \mid s)} A, \operatorname{clip}\left(\frac{\pi_{\phi}(a \mid s)}{\pi_{\phi_{\text{old}}}(a \mid s)}, 1-\epsilon, 1+\epsilon\right) A\right)$
}
\]
                    \STATE Update $\varphi$ with gradient descent using 
\[
\scalebox{0.9}{
    $\nabla_\varphi \frac{1}{|D|} \sum_{D}^{} [Q_{\varphi} - {r}_{t}+\lambda b_\theta\left(s_t,a_t\right)+\gamma \max _{a^\prime} Q_{\varphi^{\prime}}\left(s_{t+1}, a^\prime\right)  ]$
}
\]
                    \STATE Update $\theta$ using $\mathrm{Eq.}$ \ref{loss}
                \ENDFOR
                \STATE Update Q-function target networks with $\varphi' = (1-\rho)\varphi' + \rho\varphi$
        \end{algorithmic}
    \end{algorithm}

\end{document}